\documentclass[9.5pt,journal,compsoc]{IEEEtran}
\IEEEoverridecommandlockouts
\usepackage{cite}
\usepackage[colorlinks=true,linkcolor=cyan,citecolor=blue]{hyperref}
\usepackage{amsmath,amssymb,amsfonts}
\usepackage{amsthm}
\usepackage{algorithmic}
\usepackage{graphicx}
\usepackage{textcomp}
\usepackage{dsfont}
\usepackage{xcolor}
\usepackage{float}
\newtheorem{theorem}{Theorem}
\newtheorem{rem}{Remark}
\newtheorem{lemma}{Lemma}
\newtheorem{cor}{Corollary}
\newtheorem{prop}{Proposition}
\newtheorem{defn}{Definition}
\newtheorem*{Remark*}{Remark}
\def\BibTeX{{\rm B\kern-.05em{\sc i\kern-.025em b}\kern-.08em
    T\kern-.1667em\lower.7ex\hbox{E}\kern-.125emX}}
    
    \makeatletter
\newcommand{\linebreakand}{%
  \end{@IEEEauthorhalign}
  \hfill\mbox{}\par
  \mbox{}\hfill\begin{@IEEEauthorhalign}
}

\newcommand{\tr}{\text{tr}}

\makeatother
\begin{document}

\title{Clustered Graph Matching for Label Recovery and Graph Classification}

\author{Zhirui Li, Jes\'us Arroyo, Konstantinos Pantazis, Vince Lyzinski

\markboth{Journal of \LaTeX\ Class Files,~Vol.~14, No.~8, August~2015}%
{Shell \MakeLowercase{\textit{et al.}}: Bare Demo of IEEEtran.cls for IEEE Journals}

\IEEEcompsocitemizethanks{\IEEEcompsocthanksitem Z. Li is with the Department of Mathematics, University of Maryland, College Park, MD. E-mail: \href{mailto:zli198@umd.edu}{zli198@umd.edu}.
\IEEEcompsocthanksitem J. Arroyo is with the Department of Statistics, Texas A\&M University, College Station, TX. E-mail: \href{mailto:jarroyo@tamu.edu}{jarroyo@tamu.edu}.
\IEEEcompsocthanksitem K. Pantazis is with the Department of Applied Mathematics and Statistics, Johns Hopkins University, Baltimore, MD. E-mail: \href{mailto:kpantaz1@jhu.edu}{kpantaz1@jhu.edu}.
\IEEEcompsocthanksitem V. Lyzinski is with the Department of Mathematics, University of Maryland, College Park, MD. E-mail: \href{mailto:vlyzinsk@umd.edu}{vlyzinsk@umd.edu}.

\IEEEcompsocthanksitem This material is based on research sponsored by the Air Force Research
Laboratory (AFRL) and Defense Advanced Research Projects Agency (DARPA) under agreement number
FA8750-20-2-1001. 
The U.S. Government is authorized to reproduce and distribute reprints for Governmental purposes
notwithstanding any copyright notation thereon. The views and conclusions contained herein are
those of the authors and should not be interpreted as necessarily representing the official policies
or endorsements, either expressed or implied, of the AFRL and DARPA or
the U.S. Government.}}

\IEEEtitleabstractindextext{
\begin{abstract}
Given a collection of vertex-aligned networks and an additional label-shuffled network, we propose procedures for leveraging the signal in the vertex-aligned collection to recover the labels of the shuffled network. 
We consider matching the shuffled network to averages of the networks in the vertex-aligned collection at different levels of granularity.
We demonstrate both in theory and practice that if the graphs come from different network classes, then clustering the networks into classes followed by matching the new graph to cluster-averages can yield higher fidelity matching performance than matching to the global average graph. 
Moreover, by minimizing the graph matching objective function with respect to each cluster average, this approach simultaneously classifies and recovers the vertex labels for the shuffled graph. 
These theoretical developments are further reinforced via an illuminating real data experiment matching human connectomes.
\end{abstract}
}

\maketitle

\IEEEdisplaynontitleabstractindextext
\IEEEpeerreviewmaketitle

\section{Introduction}
\noindent
Graphs are powerful tools for modeling complex real-world relationships. 
A graph $G=(V,E)$ consists of two components: a set of vertices, $V$, and a set of edges, $E$, that represent connections among  the vertices. 
For instance, we can use graphs to model social networks such as Facebook or Instagram, where vertices represent single users and edges represent friendship relationships \cite{Mishra2014Social}. 
Directed graphs, which are created by adding a direction to each of its edges, can be useful to model information networks such as the World-Wide Web \cite{Broder2000WWW}. 
In epidemiology, scientists create models on a selected graph to measure and predict the spread of a certain disease, e.g., the SIR model \cite{Aldous2017SIR} and the Newman model \cite{Newman2002Disease}. 
More recent work \cite{chung2021statistical,sporns_complex} discusses the advantages of representing the brain as a graph; for example, MRI scans of patients are converted into graphs by defining neuronal regions as vertices while connections across regions are considered as edges \cite{wrg2}.
For more types of usage of graphs to model real-world complex systems, we refer the reader to  \cite{Newman2003ComplexNetwork,kolaczyk2014statistical,newman2018networks}.
Note that in the network science literature, the terms networks, nodes and links may be used in place of graphs, vertices and edges, respectively \cite{NetScienceBarabasi}; we shall use graphs/networks, vertices/nodes and edges/links interchangeably in the sequel. 

Statistical analysis of networks often begins by positing a random network model to account for the network-valued data \cite{kolaczyk2014statistical,goldenberg2010survey}.
Popular network models range from the simple Erd\H os-R\'enyi model \cite{ErdRen1963} in which all edges in the network are equally likely to exist; to the stochastic blockmodel (SBM) \cite{Holland1983} in which vertices belong to latent communities and edge probabilities depend only on the community memberships of the associated vertices; to the latent space models (LSMs) \cite{Hoff2002} in which vertices are endowed with latent positions and edge probabilities across a pair of vertices are determined by a kernel function of their associated latent positions.
These models (and their myriad variants) have conditionally independent edges (conditioned on the node memberships in SBM; conditioned on the latent positions in LSMs), a property that makes them tractable and amenable for establishing important statistical notions such as consistent estimation \cite{Bickel2009,bickel2011method,Athreya2018RDPG}, asymptotic normality \cite{athreya2013limit,tang2018limit} and  efficiency \cite{tang2017asymptotically}.
Although the simplistic nature of these models is often insufficient for capturing all the nuances of the real-world data \cite{seshadhri2020impossibility}, there is a growing literature that suggests these models can capture meaningful and important structure in even complex real networks (see, for example, \cite{wolfe_olhede_graphon,vaca2022systematic,priebe2015statistical,chung2021statistical}).

One important inference task in the network literature is that of graph matching.  
The graph matching problem seeks to find an alignment across the vertex sets of two (or more) networks that minimizes the amount of structural disagreements induced across the networks; for comprehensive surveys of the state of modern graph matching, see \cite{ConFogSanVen2004,FogPerVen2014,yan2016short}. 
In its simplest form, the \textit{graph matching problem} (GMP) is defined as follows.  
Let $\mathcal{G}_n$ be the space of undirected, loop-free, unweighted networks with $n$ vertices, and define the Frobenius norm of a matrix $X\in\mathbb{R}^{a\times b}$ as $\|X\|_F:=(\sum_{i=1}^a\sum_{j=1}^bX_{ij}^2)^{1/2}.$
Given $G_1,G_2\in\mathcal{G}_n$ with respective adjacency matrices $A$ and $B$ (so that$A_{ij}=\mathds{1}\left\{ \{i,j\}\in E(G_1)\right\},$ with $B$ defined similarly), the GMP seeks to minimize 
$\|A-PBP^T\|_F$ over all $P\in\Pi_n$, where $\Pi_n$ denotes the space of $n\times n$ permutation matrices.
Variants of the classical problem allow for the GMP to tackle weighted, directed, richly featured networks of different orders (see, for example, \cite{ModFAQ}).
Throughout this manuscript, we use the terms graph and adjacency matrix interchangeably as they provide equivalent information.

The graph matching literature is recently divided into (at least) two distinct branches: algorithmic development and theoretic graph de-anonymization (with notable cross-over work tackling provable algorithmic de-anonymization; see for example \cite{barak2019nearly,fan2020spectral}).
In the graph de-anonymization literature, a latent alignment across vertex sets is posited and the question of whether an oracle graph matching algorithm can recover this alignment under various noise models is tackled.
Recent work in this area has focused on establishing phase transitions for graph de-anonymization in terms of the error level in correlated Erd\H os-R\'enyi models \cite{cullina2016improved,cullina2017exact,Gross2011OPAN,lyzinski2014seeded,wu2021settling}, in the correlated SBM model \cite{lyzinski2016information,onaran2016optimal,racz2021correlated}, and in more general correlated edge-independent graph models \cite{lyzinski2020matchability}.
In these models, it is often assumed that edges within each network are (conditionally) independent, and that edges across the network pair are independent except that for each $\{i,j\}\in\binom{V}{2}$, $A_{ij}$ and $B_{ij}$ are positively correlated, where $\binom{V}{2}$ denotes the set of all unordered 2-tuples of distinct elements of $V$.

Inspired by the error model in \cite{arroyo2021maximum} (introduced first in the context of correlated Erd\H os-R\'enyi models in \cite{Gross2011OPAN}), we will work in the following network error model.
\begin{defn}
\label{def:bitflip}
Let $Q\in [0,1]^{n\times n}$ be a symmetric matrix.
Given ${B}\in\mathcal{G}_n$, we say that $S$ is a $Q$-errorful observation from $B$ (written $S\sim\mathrm{BF} (B,Q)$ for $S$ a ``bit-flipped'' perturbed $B$) if for each $\{i,j\}\in\binom{V}{2}$, we have that
$$S_{ij}\!=\!B_{ij}(1\!-\!X_{ij})\!+\!(1\!-\!B_{ij})X_{ij},$$
where $X_{ij}=X_{ji}\stackrel{ind.}{\sim}\mathrm{Bernoulli}(Q_{ij})$.
Note that we do not allow for self-loops in $B$ or $S$ so the diagonal elements of $Q$ are not used in this construction.
When $Q$ is the constant matrix with entries identically equal to $q$, we write $S\sim\mathrm{BF} (B,q)$ in lieu of $S\sim\mathrm{BF} (B,Q)$.
\end{defn}

\noindent This model makes no a priori assumptions on the underlying distribution of $A$, which allows for de-anonymization criteria to be established in dependent-edge network settings (i.e., in settings where edges within a network are not (conditionally) independent); see \cite{arroyo2021maximum} for detail.

\begin{rem} 
Note that in the sequel, we will be considering ``bit-flipped" perturbed graphs $S\sim\operatorname{BF}(B,Q)$ where $B$ is a Erd\H os-R\'enyi random graph with parameter $p$ (abbreviated $B\sim \mathrm{ER}(n,p)$); i.e., if
each edge is present in $B$ with probability $p$ independent of the presence or absence of all other edges.
For further connection of our ``bit-flipped" model to the graph de-anonymization phase transition work of \cite{wu2021settling, cullina2017exact} in Erd\H os-R\'enyi graphs, see Appendix \ref{app:ER}.
\end{rem}

The inference task we consider herein is a hybridization of graph matching and graph classification.
Classification tasks on networks consist of two main sub-categories: node classification and graph classification. 
Node or vertex classification considers labels at the level of vertices in the network, and seeks to use the information from a priori labeled vertices in the network to classify vertices whose label is initially unknown; note that label classification can occur within a single network or across vertices of a collection of networks.
Graph classification considers a class label at the graph level, and seeks to use the information from an a priori labeled collection of networks to classify networks whose label is initially unknown; note that graph classification must occur in the setting of multiple observed networks.
One popular method for graph-level classification is to use graph kernels to measure the similarity of graphs and then define a classifier on the similarity matrices, see \cite{Borgwardt2005KernalClass}, \cite{Nikolentzos2017MatchingNE}, \cite{shervashidze2011WLkernel}. Traditional classifiers on vectorized graphs are also equipped with regularizations that enforce some network structure \cite{vogelstein2013graph,relion2019network,wang2021learning}. Deep learning based classifiers are also popular, especially with the growing interest in neural networks; for example \cite{NiepertCNNClass,Duvenaud2015CNNClass}. Another common approach is to find a proper embedding of the graph (e.g., spectral embedding) and then build a classifier for the graphs in the embedding space; e.g., perform a hierarchical clustering via a proper metric \cite{SABARISH2020Hierarchical}.

\subsection{Shuffled Graph Classification}
\label{sec:shuff}
\noindent
The authors in \cite{vogelstein2011shuffled} consider the shuffled graph classification problem, which is the task of classifying graphs at the graph-level.
They note that when the vertex correspondences are fully observed across each pair in a collection of networks,  then classical classification methods can be used to classify graphs with unknown class types (e.g., a straight-forward classification algorithm can be implemented by choosing a suitable metric across labeled graphs and considering either the Bayes plug-in classifier or the k-nearest neighbor classifier). 
However, the paper points out that usually the assumption of fully labeled vertices is unrealistic. 
Rather, sets of shuffled graphs---which are labeled graphs with unknown labeling functions, or unlabeled graphs---should be considered instead. Under this setting, one approach is to apply a graph matching algorithm to reconcile the vertex label uncertainties, after which classical classification algorithms can be employed. 

Before defining our task further, we first remark that a pair of graphs $A$ and $B$ in $\mathcal{G}_n$ are said to be \textit{vertex-aligned} if the identity permutation is a priori known to be the true alignment across the vertex sets of the two graphs.
We note here that the notions of vertex-aligned and graph matched are subtly different.
Vertex-aligned graphs have a true, known correspondence across their vertex sets.
This alignment is often dictated by known vertex-labels or features in the network, or is provided by a subject matter expert in real data scenarios.
This true alignment is \emph{not} necessarily the optimal alignment for the graph matching problem.
This is often the case in real data networks, where the behavior of vertices across networks is not correlated as strongly as in our models (see, e.g., \cite{lyzinski2020matchability}).

Inspired by the work above, we then consider the following shuffled graph classification problem setup.
Consider a collection of $m$ vertex-aligned graphs of $k$ different classes/types (heretofore called the ``in-sample'' networks), where we model the vertex-alignment across each pair as being known a priori. 
Note that we will consider these in-sample networks as being graphs on a common vertex set.
While this could be relaxed to allow for partial alignment, the main results of the paper are analogous, and for the sake of readability, we do not pursue this further herein.
Note that if we assume that the graph class labels are initially unknown, we can estimate the class memberships of the in-sample networks via graph-level clustering.
We can then use these estimated class labels in our classification procedure.
Given an additional (``out-of-sample'') graph with both unknown type (assumed to be one of the $k$ represented in the initial collection of $m$ graphs) and unknown vertex correspondence to the collection of $m$ networks, how would we best (i) recover the vertex correspondences between the collection of in-sample networks and the out-of-sample network and (ii) classify its graph type?
Note that while we assume all graphs have the same vertex count (denoted $n$ here), this can be relaxed easily in our graph matching framework via strategic network padding; see \cite{ModFAQ}.

This is an important problem in the area of data fusion, in which two samples might come from different data sources. 
Ideally we would want to utilize all of the existing data/information (including the vertex and graph labels) in subsequent inference, and algorithms that require known vertex correspondences would require the label correspondences to be resolved across samples (e.g., tensor factorization \cite{kolda2009tensor}, joint graph embedding \cite{levin2017central,nielsen2018multiple,arroyo2021COSIE}, network regression \cite{zhou2021dynamic}, paired graph testing \cite{tang14:_semipar}, etc.).
While often we can anticipate data coming from the same source to be already matched (i.e., vertex-aligned), such assumption often would not carry over different sources. 

The main contributions of this paper are as follows:  We provide a novel exploration of the problem of matching a label-shuffled graph to a collection of vertex-aligned networks (Section \ref{sec:problem_form}).  We provide approaches for matching the shuffled graph to the matched collection at three levels of granularity: matching to a coarse average (Section \ref{sec:k2}), to a clustered average (Section \ref{sec:clusmatch}), and to each graph individually.
Throughout, we provide both theory and illustrative experiments showing the benefits/costs of matching at each level of granularity based on the latent structure of the a priori matched collection, with an emphasis on the benefit of clustered matching if the clusters in the matched collection are sufficiently different.

\noindent
\textbf{Notation:} For functions $f,g:\mathbb{Z}_{\geq0}\mapsto \mathbb{R}_{\geq0}$, we will make use of the following standard asymptotic notation: we write $f=O(g)$ if $\exists\, C> 0,$ and $n_0\in\mathbb{Z}_{\geq0}$
     s.t. $f(n)\leq C g(n)$ for $n\geq n_0$;
     $f=\Omega(g)$ if $\exists\, C> 0,$ and $n_0\in\mathbb{Z}_{\geq0}$
     s.t. $ C g(n)\leq f(n)$ for $n\geq n_0$;
    $f=\Theta(g)$ if $f=\Omega(g)$ and $f=O(g)$;
    $f=o(g)$ if $\lim_{n\rightarrow\infty}\frac{f(n)}{g(n)}= 0$;
    $f=\omega(g)$ if  $\lim_{n\rightarrow\infty}\frac{g(n)}{f(n)}= 0$.

\section{Clustered graph matching for classification}
\label{sec:problem_form}
\noindent
Before formally defining our graph matching setup, we first note that all graphs/parameters considered herein are implicitly indexed by $n$; so that the background graphs $B^{(i)}$ are graph sequence $\{B^{(i)}_n\}_n$, permutations $P$ are permutation sequence $\{P_n\}_n$, with model parameters $m=m_n$, $k=k_n$, $\xi=\xi_n$, $p=p_n$, etc., all varying in $n$. 
 In the sequel, we suppress the $n$-index moving forward to ease notation. 

Formally, the problem we consider is defined as follows.
Suppose $B^{(1)},B^{(2)},\ldots,B^{(k)}\in\mathcal{G}_n$ denote $k$ vertex-aligned, unobserved graphs; each of them represents a distinct graph type/class (in the classification framework). 
We will consider both settings in which these background graphs are assumed to be latent and fixed, or in which they are assumed to be latent graph-valued random variables.
In the latter case, we will condition on the $B^{(j)}$'s before generating the subsequent $S^{(j)}_i$'s below.
For each $j\in[k]:=\{1,2,3,\cdots,k\}$ let $m_j\in\mathbb{N}$ be such that $\sum_j m_j = m$, and consider 
$S^{(j)}_i\sim\operatorname{BF}(B^{(j)},p_j)$ for $0<p_j<1/2$, $i=1,2,\ldots,m_j$, and further assume that the collection of graphs 
$\{\{S^{(j)}_i \}_{i=1}^{m_j}\}_{j=1}^k$ are conditionally independent given $B^{(1)},B^{(2)},\ldots,B^{(k)}.$
The assumption that $p_j<1/2$ is justified as follows.
   If $p_j=1/2$ for some $j$, then $S^{(j)}_i
   \stackrel{i.i.d.}{\sim}$ER$(n,1/2)$, and they carry no information on $B^{(j)}$. 
    If $p_j>1/2$, then $\operatorname{BF}(B^{(j)},p_j)\stackrel{\mathcal{L}}{=}\operatorname{BF}(\bar{B}^{(j)},1-p_j)$, where $\bar{B}^{(j)}$ is the complement graph of $B^{(j)}$. 
    Thus, by replacing $B_j$ with its complement we can reduce to the case where $p_j<1/2$.
    
For each $j\in[k]$, the graphs in $\{S^{(j)}_i\}_{i=1}^{m_j}$ represent the observed in-sample networks of type $j$, which can be thought of as edge-noisy, vertex-aligned versions of the background graph $B^{(j)}$.
Consider further a fixed $r\in[k]$ and further simulate $A\sim \operatorname{BF}(B^{(r)}, p_r)$ independent (conditionally given the $B^{(1)},B^{(2)},\ldots,B^{(k)}$) 
of all $\{S^{(j)}_i\}$ where $0<p_r<1/2$ are fixed; letting $P^*$ be a fixed but unknown permutation in $\Pi_n$, we observe $R=(P^*)^TAP^*$, which here represents the out-of-sample,  label-obfuscated graph.

Our task then is as follows:
given the collection of vertex-aligned networks $\{S^{(j)}_i\}$, we seek to recover both the vertex alignment (here $P^*$) and the graph label (here $h$) of $R$.
Matching $R$ to $\{S^{(j)}_i\}$ to recover the correct vertex alignment of $R$ can here proceed at (at least) three levels of granularity:
\begin{itemize}
    \item[i.]  (Coarse matching) Define the global average matrix $C$ by $C = \frac{1}{m}\sum_{i,j} S^{(j)}_i$; note each entry of $C$ is in the interval $[0,1]$.  We can match $R$ to $C$ to recover the labels of $R$.
    \item[ii.]  (Clustered matching)
     Compute the class-level graph means: Let $\ell\in[k]$, and let $\mathcal{C}_\ell$ be the set of graphs in class $\ell$, define 
    $$C_\ell:=\frac{1}{|\mathcal{C}_\ell|}\sum_{S^{(j)}_i\in \mathcal{C}_\ell }S^{(j)}_i.$$
    Match $R$ to each $C_\ell$, computing $\Delta_\ell=\min_{P\in\Pi_n}\|C_\ell-PRP^T\|_F.$
    Letting $\ell^*\in\text{argmin}_\ell\Delta_\ell$, classify $R$ as type $\ell^*$ and label $R$ via $P_{\ell^*}\in \text{argmin}_{P\in\Pi_n}\|C_{\ell^*}-PRP^T\|_F$.
    Note that if the class labels are initially unobserved for the in-sample graphs, we can obtain estimated labels via clustering the $S^{(j)}_i$'s into $k$ clusters, and then use these cluster assignments as class labels for the above procedure.
    \item[iii.] (Fine matching)     Match $R$ to each $S^{(j)}_i$, computing $$\Delta_{ij}=\min_{P\in\Pi_n}\|S^{(j)}_i-PRP^T\|_F.$$
    Letting $\{i^*j^*\}\in\text{argmin}_{ij}\Delta_{ij}$, label $R$ via $P_{\{i^*j^*\}}\in \text{argmin}_{P\in\Pi_n}\|S^{(j^*)}_{i^*}-PRP^T\|_F$.
\end{itemize}

While we suspect (and empirically it is often the case; see Section \ref{sec:hnu1}) that the clustered matching strategy would yield the highest fidelity recovery of $P^*$ (i.e., of the permutation that unshuffles $R$), this is not always the case.
Indeed, the data smoothing obtained via cluster/class averaging can yield worse matchings if there is sufficient variability/bias across the elements being averaged, in which case the fine matching may yield higher fidelity results.
While this is an important issue to untangle, we do not pursue this further here as in our simulations and experiments, clustered averaging yields the best (or close to the best) results.

There is a further computational advantage to clustered matching, as it only requires computing $m$ matchings.  
In settings where $m$ and $n$ are large, computing all pairwise matchings can be prohibitively expensive.
At the other extreme, while coarse matching is computationally less expensive, if there is significant structural differences across $B^{(1)},B^{(2)},\ldots,B^{(k)}$, then it is natural to expect the signal of the true cluster to be whitened out in $C$, and matching $R$ to $C$ will not recover  $P^*$.
We shall demonstrate below that the clustered matching balances computational feasibility and within-class signal fidelity to produce an accurate, more scalable estimate of  $P^*$.
Moreover, the clustered matching alone is able to solve both aspects of our inference task simultaneously, both matching and classifying $R$ in one step.

\section{The good and the bad of coarse matching}
\label{sec:k2}

In this section, we explore both the potential benefits and potential problems associated with the coarse matching strategy.
We consider first the case where $k=2$; i.e., where we have two distinct asymmetric (i.e., 
$PB^{(i)}P^T\neq B^{(i)}$ for all $P\neq I_n$) background graphs $B^{(1)}$ and $B^{(2)}$.
Suppose further 
\begin{equation}
    \label{eq:key}
    \{I_n\}\notin\text{argmin}_{P\in\Pi_n}\|B^{(1)}-PB^{(2)}P^T\|_F.
\end{equation}
Note that if Eq. \ref{eq:key} did not hold, then (under modest assumptions) coarse matching would be successful in unshuffling $R$ with high probability.
Eq. \ref{eq:key} is necessary for us to explore the break-down point when coarse matching may fail and clustered matching succeed.
We note here that in this section $B^{(1)}$ and $B^{(2)}$ are still modeled as vertex-aligned in that the true underlying permutation between graphs is still the identity matrix. 
 The setting in this section captures the often-true reality that the true underlying permutation (according to the assigned data labels) is not Graph Matching optimal.

Without loss of generality, let $A\sim \mathrm{BF}(B^{(1)},p_1)$ for $p_1\in(0,1/2)$, so that $R=(P^*)^TAP^*$ is our out-of-sample network and $P^\ast$ is the correct permutation that unshuffles $R$.
Here, matching $R$ to $C$ amounts to trying to find $P^\ast$ by solving the following quadratic assignment problem (where, to ease notation, $f(P):=\sum_{ij}\text{tr}(S_i^{(j)}PRP^T)$):
$$
\min_{P}\|C-PRP^T\|_F\!\Leftrightarrow\!
\max_P \sum_{ij}\text{tr}(S_i^{(j)}PRP^T)\!\Leftrightarrow\!
\max_P f(P)
$$
Letting $\mathbb{E}_B(\cdot)=\mathbb{E}(\cdot|B^{(1)},B^{(2)})$, if $\bar B^{(1)}\in\mathcal{G}_n$ (resp., $\bar B^{(2)}$) denotes the complement graph of $B^{(1)}$ (resp., $B^{(2)}$) we have, where $P^\bigstar:=P(P^*)^T$ to ease notation, 
\begin{align}
\label{eq:needfor5}
\mathbb{E}_B&(\text{tr}(S_i^{(j)}PRP^T))
\!=\! (1\!-\!p_j)(1\!-\!p_1)\text{tr}(B^{(j)} P^\bigstar B^{(1)} (P^\bigstar)^T) \notag\\
&+ p_j(1-p_1)\text{tr}(\bar B^{(j)} P^\bigstar B^{(1)} (P^\bigstar)^T) \notag\\
&+ (1-p_j)p_1\text{tr}(B^{(j)} P^\bigstar \bar B^{(1)} (P^\bigstar)^T)\notag\\
&+ p_jp_1\text{tr}(\bar B^{(j)} P^\bigstar \bar B^{(1)} (P^\bigstar)^T)\\
=&  (1\!-\!2p_j\!)(1\!-\!2p_1)\text{tr}(B^{(j)} P^\bigstar\! B^{(1)} (P^\bigstar)^T)\!+\!T(B^{(1)}\!,B^{(j)}\!,n).\notag
\end{align}
where $T(B^{(1)},B^{(j)},n)$ is independent of $P$ and $P^*$ (see Appendix \ref{app:eq7}).

\subsection{The benefits of averaging}
\label{sec:benefit}
\noindent
Assume for the moment that $P^\bigstar:=P(P^*)^T$ shuffles exactly $\xi$ labels, 
and that 
$\mathbb{E}_B(f(P)-f(P^*))<0$,
which implies that $P^\ast$ is better than $P$ for matching 
$R$ to $C$, on average. 
This condition ensures that there are enough ``good" matches to $A$ (i.e., those from the same background) in the in-sample set to mitigate the effect of averaging the entire collection of $m$ networks, as those from $B^{(2)}$ will, with high probability, not match correctly to $A$.
\begin{prop}
\label{prop:33}
With notation as above, if 
\begin{align}
    \label{eq:growth}
    -\mathbb{E}_B(f(P)-f(P^*))=\omega(m {\xi}\sqrt{n\log n}),
\end{align}
holds for all {$\xi\in\{2,3,\cdots,n\}$} and all $P$ such that $P(P^*)^T\in \Pi_{n,\xi}$ (where $\Pi_{n,\xi}$ is the set of permutations shuffling exactly {$\xi$} labels),
then  \begin{align*}
        \mathbb{P}(\{&P^*\}\notin \text{argmin}_P\|C-PRP^T\|_F)=e^{-\omega(\log n)}  
    \end{align*}
\end{prop}
\noindent The proof of this proposition combines McDiarmid's inequality with a standard union over such $P$ and $\xi$; see Appendix \ref{app:addcomp} for the derivation).
Note that this union bound combined with McDiarmid or similar concentration bounds  is a standard argument in the literature, appearing in multiple other graph matching works 
(see, for example, \cite{lyz2016relax,sussman2018matched,lyzinski2020matchability} among others).
Lastly, as an example of the feasibility of Eq. \ref{eq:growth}, note that if the background graphs $B^{(i)}\sim$ER$(n,q_i)$, then under mild assumptions 
$-\mathbb{E}_B(f(P)-f(P^*))=\Theta(m \xi n)\in \omega (m \xi\sqrt{n\log n})$, and Eq. \ref{eq:growth} holds.

\subsection{The cost of averaging}
\label{sec:cost}
\noindent
The case where coarse averaging is detrimental to matchability is a bit more nuanced.
Assume that there exists a $P$ such that $P(P^*)^T\in\Pi_{n,\xi}$, and $\mathbb{E}_B(f(P)\!-\!f(P^*))\!>\!0$.
    This is tantamount to the noise contributed by the class 2 graphs obfuscating the alignment  signal present in the in-sample class 1 graphs.
    Indeed, the optimal graph matching permutation between a class 1 and class 2 graph will, with high probability, not be the true latent (in the case of the out-of-sample graph) or observed (in the case of in-sample graphs) alignment.
    
    To see the effect of averaging with this noise, we first define 
    for each $x\in\{0,1\}^4$, the   following quantity, which captures the edge/non-edge patterns in the graphs before and after shuffling,
\begin{align*}
    N_x:=\bigg|&\bigg\{\,\{h,\ell\}\in\binom{V}{2}\text{ s.t. }\bigg(B^{(1)}[\sigma(h),\sigma(l)],\\
    &B^{(1)}[h,l],B^{(2)}[\sigma(h),\sigma(l)],B^{(2)}[h,l]\bigg)=x\bigg\}\bigg|.
\end{align*}

    We then have the following theorem (see Appendix \ref{app:pf1} for the proof using Stein's method).
    \begin{theorem}
    \label{thm:thm1}
    Under the setup as above, let $p_1=p_2=p$ for fixed $p\in(0,1/2)$. 
    If any of the following conditions hold
    \begin{itemize}
    \item[i.]$|m_1-m_2|=o(m)$ and 
  $N_{1110}+N_{0001}=\omega((n\xi)^{2/3});$
%
\item[ii.]$m_1,m_2=\Theta(m),\,|m_1-m_2|=\Theta(m)$ and 
      $N_{1110}+N_{0001}+N_{1001}+N_{0110}=\omega((n\xi)^{2/3});$
    \item[iii.]$m_2/m_1=\omega(1)$ and
      $N_{1110}+N_{0001}+N_{1001}+N_{0110}=\omega((n\xi)^{2/3});$
     \item[iv.] $\frac{n\xi}{m^3}=\omega(1)$,
\end{itemize}
then we have that  $$\frac{f(P)-f(P^*)-\mathbb{E}_B(f(P)-f(P^*))}{\sqrt{\text{Var}_B(f(P)-f(P^*))}}$$ converges in law to a standard normal random variable
    with 
    $$\text{Var}_B(f(P)-f(P^*))=O(n\xi m^2).$$
    \end{theorem}
\noindent 
    The conditions in Theorem \ref{thm:thm1} ensure that $B^{(1)}$ and $B^{(2)}$ have sufficiently many edgewise structural differences post-shuffling to provide an adequate sample size for Stein's normal approximation method to provide approximate normality of $f(P)-f(P^*)$, as well as sufficient variance growth for $f(P)-f(P^*)$ which will be used later to provide sharp concentration of this difference.
We suspect these precise conditions are not necessary, and can be relaxed with more careful analysis of the mismatch between $B^{(1)}$ and $B^{(2)}$, though we do not pursue this further here. 
    
As an immediate consequence of Theorem \ref{thm:thm1}, we have the following corollary, which shows that the incorrect permutation $P$ is a better solution of the quadratic assignment problem.
\begin{cor}
\label{cor:cor1}
Given the conditions of Theorem \ref{thm:thm1}, if $\mathbb{E}_B(f(P)-f(P^*))>0$ we have the following:
    \begin{itemize}
        \item[i.] With no further assumptions on $\mathbb{E}_B(f(P)-f(P^*))$, we have that $\mathbb{P}(f(P)>f(P^*))\geq 1/2(1-o(1)).$
        \item[ii.] If we assume that  $\mathbb{E}_B(f(P)-f(P^*))=\omega(m\sqrt{n\xi\log n}),$ we have that $\mathbb{P}(f(P)>f(P^*))\geq 1-o(1).$
    \end{itemize}
    \end{cor}  
\noindent Note that in the case where $\mathbb{E}_B(f(P)-f(P^*))<0$ for every $P\neq P^*$, if we do not provide an associated growth rate, then the same proof as in Theorem \ref{thm:thm1} yields $\mathbb{P}(f(P)>f(P^*))\leq 1/2(1-o(1))$. The growth rate assumption in Eq. \ref{eq:growth} is made to uniformly bound these probabilities close to $0$.
\begin{rem}
 Sections \ref{sec:benefit} and \ref{sec:cost} imply that the key for correctly recovering the latent vertex alignment for the out-of-sample graph is 
    $$\frac{
    \text{tr}(B^{(2)}B^{(1)})-\text{tr}(B^{(2)} P^\bigstar B^{(1)} (P^\bigstar)^T)
    }{
    \text{tr}(B^{(1)} P^\bigstar B^{(1)} (P^\bigstar)^T)\!-\!\text{tr}(B^{(1)}B^{(1)})} <\frac{m_1(1-2p_1)}{m_2(1-2p_2)}$$
  This is akin to a signal--to--noise ratio bound, so that coarse matching is successful if the noise contributed by the class 2 graphs is comparatively small.
  Note that there is a gap in the growth rates of Eq. \ref{eq:growth} and Corollary \ref{cor:cor1} used to ensure coarse matching will/will not fail with high probability. 
  While we suspect a sharp phase transition is present, we do not pursue this further herein.
\end{rem} 

    \subsection{Matching when k greater than 2}
    \label{sec:kb2}
    \noindent
    We next consider cases where $k>2$; i.e., where we have multiple distinct backgrounds $B^{(1)},B^{(2)},\ldots, B^{(k)}$.
    Further suppose that for all $j = 2,3,\ldots,k$,
    \begin{equation*}
    \{I_n\}\notin\text{argmin}_{P\in\Pi_n}\|B^{(1)}-PB^{(j)}P^T\|_F.
    \end{equation*}
    Without loss of generality, let $A\sim \mathrm{BF}(B^{(1)},p_1)$, so that we observe $R=(P^*)^TAP^*$.
    In the $k=2$ case, we saw that the noise contributed by the graphs not from the background class of $A$ (i.e., the one satisfying Eq. \ref{eq:key}) could overwhelm the signal provided by the graphs from the same background class as $A$.
    When $k>2$, the effect of this noise can be more nuanced. 

    In one direction, note that
the same McDiarmid's inequality argument as in the $k=2$ case yields 
that if 
    $
    -\mathbb{E}_B(f(P)-f(P^*))
    $
    is sufficiently big for all $P\neq P^*$, then, with high probability, matching $R$ to $C$ will yield the correct alignment.
    
In the other direction, if there exists a $P$ such that $P(P^*)^T$ shuffles {$\xi$} vertex labels and $\mathbb{E}_B(f(P)-f(P^*))>0$,
    then we have the following result (which is an immediate corollary of the analogue of Theorem \ref{thm:thm1} in the present setting).
   \begin{cor}
Under the setup as above with $p_i=p$ for all $i\in[k]$, if $n{\xi}/m^3=\omega(1)$ and $\mathbb{E}_B(f(P)-f(P^*))>0$, then
    \begin{itemize}
        \item[i.] with no further assumptions on $\mathbb{E}_B(f(P)-f(P^*))$, we have that $\mathbb{P}(f(P)>f(P^*))\geq 1/2(1-o(1)).$
        \item[ii.] if we assume that  $$\mathbb{E}_B(f(P)-f(P^*))=\omega(m\sqrt{n\xi\log n}),$$ we have that $\mathbb{P}(f(P)>f(P^*))\geq 1-o(1).$
    \end{itemize}
   \end{cor}

    \noindent As in the $k=2$ case, the behavior hinges on $\mathbb{E}_B(f(P)-f(P^*))$, which can be more nuanced in the $k>2$ setting, as the following example illuminates.
    
    For each $i=1,2,3$, let $B^{(i)}\!\stackrel{ind.}{\sim}\!\operatorname{SBM}(3n, [n,n,n], \Lambda^{(i)})$, so that for each $i$, the $3n$ vertices in $B^{(i)}$ are divided into three communities, each of size $n$. Next, sample independent $S^{(i)}_{h}\sim\mathrm{BF}(B^{(i)},p_i)$  and let $A\sim\mathrm{BF}(B^{(1)},p)$. 
Then, 
\begin{align*}
    \mathbb{E}\text{tr}(P^TAPC)=&\frac{m_1}{m}\mathbb{E}\text{tr}(P^TAPS^{(1)}_{1})
    +\frac{m_2}{m}\mathbb{E}\text{tr}(P^TAPS^{(2)}_{1})\\
    &+\frac{m_3}{m}\mathbb{E}\text{tr}(P^TAPS^{(3)}_{1}).
\end{align*}
 Let $b_i:V\mapsto \{1,2,3\}$ denote the community membership function (so that $b_i(v)=j$ if vertex $v$ is in community $j$), and assume that $b_1=b_2=b_3$ with
    $$b_i(v)=
    1+
    \mathds{1}\{n+1\leq v\leq 2n\}+
    2*\mathds{1}\{2n+1\leq v\leq 3n\}.$$
    For each $i$, $\Lambda^{(i)}\in[0,1]^{3\times 3}$ is a symmetric $3\times 3$ matrix such that
for each $\{u,v\}\in\binom{V}{2}$, (where $E_i$ is the set of edges of $B^{(i)}$),
$\mathds{1}\{\{u,v\}\in E_i\}\stackrel{ind.}{\sim}\operatorname{Bernoulli}(\Lambda^{(i)}[b_i(u),b_i(v)]).$

Let $p_1=p$, $p_2=p_3=q$, and let $a>r>0$, and $\epsilon>0$. 
Define $\Lambda^{(i)}$ as follows.
Each $\Lambda^{(i)}$ has all entries identically equal to $r$ except that
$\Lambda^{(1)}[1,1]=a$,
$\Lambda^{(2)}[2,2]=a+\epsilon,$ and
$\Lambda^{(3)}[3,4]=a+\epsilon.
$
To demonstrate the complications of averaging multiple backgrounds, consider for example (among other similar choices)
$a=0.3,\,\epsilon=0.5;\, r=0.1,\, p=0.4,\,q=0.1.$
Let $P$ be any fixed permutation that flips all the vertices between blocks 1 and 2. When $m_2=2m_1$ and $m_3=0$,  
\begin{align*}
    \mathbb{E}\text{tr}(AC)\!=\!c_1 n^2(1\!-\!o(1));\,\,
    \mathbb{E}\text{tr}(P^T\!APC)\!=\!c_2n^2(1\!-\!o(1));
\end{align*}
and when $m_2=m_1=m_3$, 
\begin{align*}
    \mathbb{E}\text{tr}(AC)\!=\!c_1 n^2(1\!-\!o(1));\,\,
    \mathbb{E}\text{tr}(P^T\!APC)\!=\!c_3 n^2(1\!-\!o(1)).
\end{align*}
where $1<c_3<c_1<c_2<2$ are constants that can be obtained from direct mathematical computation.
As $\text{tr}(AC)$ and $\text{tr}(P^TAPC)$ concentrate tightly about their means, we see that for sufficiently large $n$, flipping blocks 1 and 2 via $P$ (an optimal alignment of $\Lambda^{(1)}$ and $\Lambda^{(2)}$) when $m_2=2m_1$ and $m_3=0$ will, with high probability, result in a better match for the average than the true identity alignment.  
This is unsurprising, as $\Lambda^{(2)}$ is designed for this end; i.e., to attract the dense block in $\Lambda^{(1)}$ to block 2 in $\Lambda^{(2)}$.
If, however, the wrong-class in-sample graphs are evenly split between classes 2 and 3 with $m_1$ from each of the three classes, 
then the alignment provided by $P$ is no longer better than the identity alignment (again with high probability).
Noting the same analysis holds for flipping blocks 1 and 3 (an optimal alignment of $\Lambda^{(1)}$ and $\Lambda^{(3)}$), we see here that the noise from the in-sample, wrong-class networks effectively cancels across classes as the wrong-classes pushing the optimal permutation in different, counteracting directions.

It is clear that if all the $P^{(i)}$'s that optimally align $B^{(1)}$ and $B^{(i)}$ are equal (or overlap significantly), then the noise cancellation demonstrated in the example above will not occur.
In the SBM setting, this can be achieved by ensuring that the optimal alignment of $\Lambda^{(i)}$ to $\Lambda^{(j)}$ is the identity mapping for $i,j\neq 1$. 
We next seek to generalize this idea to other network models.
To this end, we consider the following multiple random dot product graph model from \cite{arroyo2021COSIE}.
    \begin{defn}
    \label{def:cosie}
    Let $U$ be an $n\times d$ matrix with orthonormal columns, and for $j\in[n]$, let $U_j$ denote the $j$-th row of $U$.
    Let $R^{(1)}, \ldots, R^{(m)}$ be $d\times d$ symmetric matrices such that $0 \leq$ $U_{j} R^{(i)} U_{h}^T \leq 1$ for all $j,h \in[n],\, i \in[m]$. 
    We say that the random adjacency matrices $B^{(1)}, \ldots, B^{(m)}$ are jointly distributed according to the common subspace independent-edge graph (COSIE) model with rank $d$ and parameters $U$ and $R^{(1)}, \ldots, R^{(m)}$ if given $U$ and $\{R^{(i)}\}_{i=1}^m$, the collection of networks $\{B^{(i)}\}_{i=1}^m$ is independent, and for each $i\in[m]$, the upper-triangular entries of  $B^{(i)}$ are independent and distributed according to
    \begin{align*}
        &\mathbb{P}(B^{(i)} \!\mid\! U, R^{(i)})\\
        &=\prod_{j<h}\left(U_{j} R^{(i)} U_{h}^T\right)^{B^{(i)}[j,h]}\left(1-U_{j} R^{(i)} U_{h}^T\right)^{1-B^{(i)}[j,h]}
    \end{align*}
    \end{defn}
    \noindent The COSIE model of Definition \ref{def:cosie} provides a flexible framework for modeling a collection of networks on a common vertex set, and it encompasses many important network models including the multilayer stochastic blockmodel of \cite{Holland1983}.
    The score matrices $R^{(i)}$ in the COSIE model allow us a similar opportunity as in the SBM setting to ensure that the wrong-class, in-sample graphs are all misaligned in synchrony.
    We shall now demonstrate this in the following example.
    
 Assume that $B^{(1)}, \ldots, B^{(m)}$ are jointly distributed according to the COSIE model with rank $d$ and parameters $U$, $R^{(1)}, \ldots, R^{(m)}$, and assume further that the $R^{(j)}$'s are diagonal matrices for all $j$ (this is similar to the model considered in \cite{wang2019joint,draves2020bias}).
    Suppose further that the diagonal of $R^{(1)}$ are ordered to be non-decreasing, and that there exists a common $Q\in\Pi_d\setminus\{I_d\}$ such that for all $j\in[m]\setminus\{1\}$,
    \begin{align*}
        Q&\in\operatorname{argmin}_{P\in\Pi_d}\|R^{(1)}-PR^{(j)}P^T\|_F,\\
        I_d&\notin\operatorname{argmin}_{P\in\Pi_d}\|R^{(1)}-PR^{(j)}P^T\|_F.
    \end{align*}
    The following lemma, proven in Appendix \ref{app:pf2}, will codify sufficient conditions under which wrong-class in-sample graphs are all misaligned in synchrony.
    \begin{lemma}
    \label{prop:Q}
    With setup as above, if there exists a permutation $P\in\Pi_n$ such that for all $j\neq 1$,
    \begin{align}
    \label{eq:tech}
    1-2\|U^T P U-Q\|_F
    >\frac{\operatorname{tr}(R^{(1)}I_d R^{(j)}I_d^T)}{\operatorname{tr}(R^{(1)}QR^{(j)}Q^T)},
    \end{align}
    then
   $\forall\,j\!\neq \!1$,
          $\operatorname{tr}(P^T \mathbb{E}(B^{(1)})P\mathbb{E}(B^{(j)})\!>\!\operatorname{tr}( \mathbb{E}(B^{(1)})\mathbb{E}(B^{(j)})).$
    \end{lemma}
    \noindent The technical Lemma condition (Eq. \ref{eq:tech}) is used to ensure that the action of shuffling $\mathbb{E}(B)$ by $P$ (and yielding $PUR^{(i)}U^TP^T$) is sufficiently close to the action of shuffling $R^{(i)}$ by $Q$ (and yielding $UQR^{(i)}Q^TU^T)$; this is used to then lift the shuffling of the unknown $R^{(i)}$'s to a shuffling of the observed $B^{(i)}$'s.
  
    Next, define
    $$
    \tilde f_j(P):=\operatorname{tr}(P^T \mathbb{E}(B^{(1)})P\mathbb{E}(B^{(j)})-\operatorname{tr}( \mathbb{E}(B^{(1)})\mathbb{E}(B^{(j)})).
    $$
    If $P$ satisfies the conditions in Lemma \ref{prop:Q} and  $\sum_{i\neq 1}m_i$ is sufficiently large relative to $m_1$, we have
    \begin{equation}
    \label{eq:fp}
        \sum_{i\neq 1} m_i\tilde f_i(P)> -m_1\tilde f_1(P).
    \end{equation}
    Consider now the setting where $p_1=\cdots=p_k=p$, and let $A\sim \mathrm{BF}(B^{(1)},p)$ and let
    $\{S^{(i)}_j\}$ as before.
    We seek then to match the observed network $R=(P^*)^T AP^*$ with $C=\frac{1}{m}\sum_{i,j}S^{(i)}_j$.
    Eq. \eqref{eq:fp} ensures that 
    \begin{align*}
     \mathbb{E}&\left(\operatorname{tr} \left( P^T A P C\right)\right)=\mathbb{E}\left(\mathbb{E}_B\left(\operatorname{tr} (P^T A  P C)\right)\right)\notag\\
 &=\sum_i \frac{m_i(1-2p)^2}{m}\operatorname{tr} (P^T \mathbb{E}(B^{(1)})  P \mathbb{E}(B^{(i)}))\notag\\
 &>\sum_i \frac{m_i(1-2p)^2}{m}\operatorname{tr} ( \mathbb{E}(B^{(1)}) \mathbb{E}(B^{(i)}))\notag\\
    &=\mathbb{E}\left(\mathbb{E}_B\left(\operatorname{tr} ( A  C)\right)\right)=\mathbb{E}\left(\operatorname{tr} \left(A C\right)\right).
   \end{align*}
A similar application of Stein's method as in Theorem \ref{thm:stein} will yield that
$
\operatorname{tr}( A(PCP^T-C))
$
suitably scaled and centered will converge to a standard normal random variable.
This will yield the following theorem.
\begin{theorem}
\label{thm:cosie}
With assumptions as in Lemma \ref{prop:Q}, assume that $p_i=p$ for some fixed $0<p<1/2$ for all $i\in[k]$. 
Letting $P$ satisfy the conditions of Lemma \ref{prop:Q}, and assume that $\{m_i\}$ is such that Eq.\@ \ref{eq:fp} holds.
If $P(P^*)^T$ shuffles $\ell$ vertex labels, then 
$n\ell/m^3=\omega(1)$ implies that
    \begin{itemize}
        \item[i.] with no further assumptions on $\mathbb{E}\operatorname{tr}( A(PCP^T-C))$, we have that 
$\mathbb{P}(f(P)>f(P^*))\geq 1/2(1-o(1)).$
        \item[ii.] if we assume that  $\mathbb{E}\operatorname{tr}( A(PCP^T-C))=\omega(\sqrt{n\ell\log n}),$ we have that
 $\mathbb{P}(f(P)>f(P^*))\geq 1-o(1).$
    \end{itemize}
\end{theorem}

\section{Clustered matching}
\label{sec:clusmatch}

    \begin{figure*}[t!]
    \centering
    \includegraphics[width =\textwidth]{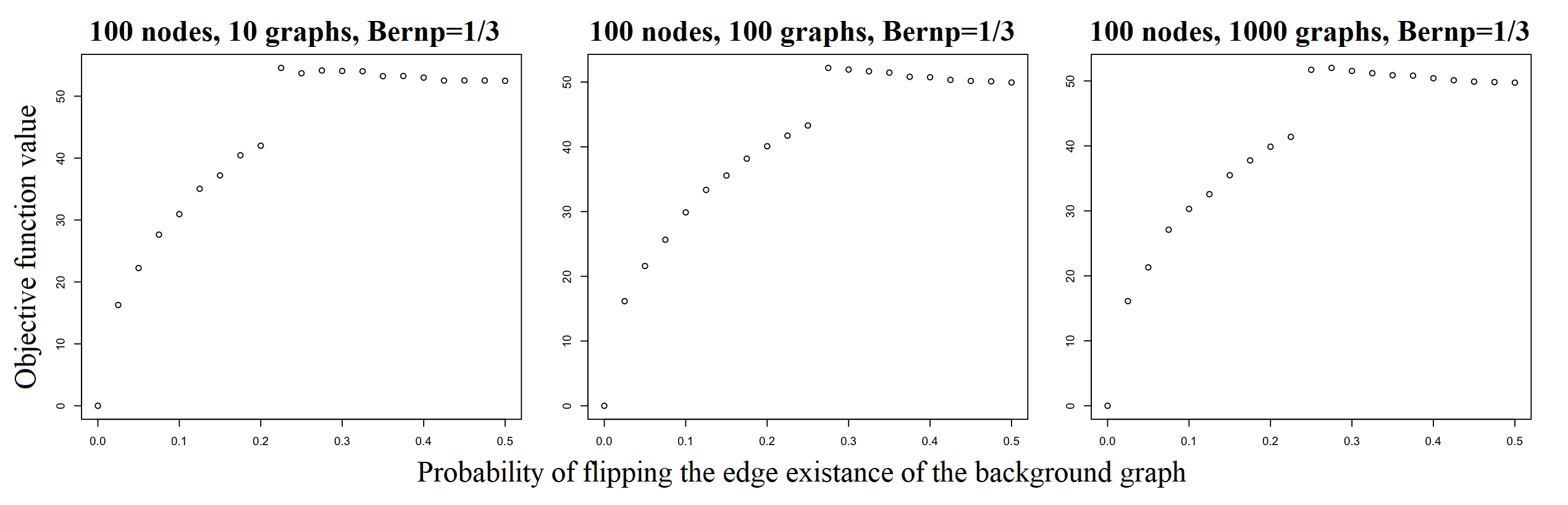}
    \vspace{-5mm} 
    \caption{With a single background $B\sim\operatorname{ER}(n=100,1/3)$, we consider $A,S^{(1)}_i\stackrel{i.i.d.}{\sim}$BF$(B,q)$, and we match $A$ (i.e., $P^*=I_n$) to $C$
    using \texttt{SGM} with 5 seeds.
    Varying the number of in-sample graphs ($m=10$ in the left panels, $m=100$ in the middle panels, and $m=1000$ in the right panels), we plot the \texttt{SGM} objective function value $f=\|A-PCP^T\|_F$ versus the value of the edge perturbation parameter $q$, averaged over 10 Monte Carlo iterates.}
    \label{fig:er2}
    \end{figure*}
    \noindent
Consider next the case of clustered matching, where for simplicity we will assume the class labels are observed or the clustering perfectly recovers the class labels amongst the in-sample networks $S^{(i)}$.
The case in which the clusters are noisily recovered is of great interest, and will be the subject of subsequent work.
For each $i\in[k]$, let $C^{(i)}$ be the cluster average of the graphs from class $i$, so that
$
C^{(i)}=\frac{1}{m_i}\sum_{j=1}^{m_i}S^{(i)}_j.
$
With $A\sim\operatorname{BF}(B^{(1)},p_1)$ as before (recall that we observe the shuffled $A$, i.e., $R=(P^*)^TAP^*)$, recalling the form of $
    \mathbb{E}_B(\text{tr}(C^{(i)}PRP^T))$ from Eq. (\ref{eq:needfor5}), we have that for $P\neq P^*$ and $p=p_1=\cdots=p_k$ (note that this equation is shown in Appendix \ref{app:eq7}),
\begin{align}
    \mathbb{E}_B&(\text{tr}(C^{(1)}P^*R(P^*)^T))-
    \mathbb{E}_B(\text{tr}(C^{(i)}PRP^T))
    \notag\\
    &=(1-p)(1-2p)\|B^{(1)}\|_F^2-p(1-2p)\|B^{(i)}\|_F^2
    \notag\\
    &\hspace{5mm}-
    (1-2p)^2\text{tr}(B^{(i)}P(P^*)^TB^{(1)}P^*P^T)
    . \label{eq:cmg2}
\end{align}
\begin{theorem}
\label{thm:secIV}
With notation as above,  denote $X_{i,P}\!=\!\text{tr}(C^{(1)}P^*R(P^*)^T)\!-\!
    \text{tr}(C^{(i)}PRP^T).$
    If for all integer $2\leq \xi\leq n$, and for all $i\neq 1$, and $P$ s.t. $P(P^*)^T\in\Pi_{n,\xi}$, we have
Eq. \ref{eq:cmg2} is of order $\omega(n\sqrt{\xi\log(n)})$
, then
\begin{align}
\label{eq:71}
    \mathbb{P}_B(\exists i\in[k]\setminus\{1\}, P\in\Pi_n\text{ s.t. }X_{i,P}\leq 0)\!=\!e^{-\omega(\log(n))},
\end{align}
\end{theorem}
\noindent The proof of Theorem \ref{thm:secIV} is a straightforward application of Hoeffding's inequality; see Appendix \ref{app:eq71} for detail.

Theorem \ref{thm:secIV} implies that with high probability the correct matching of $R$ to $C^{(1)}$ will yield a better objective function value than any other matching of $R$ to any other class mean.
Hence, clustered matching can be used to both unshuffle \textit{and} classify $R$ by assigning it to the cluster/class it matches best to (best as in lowest objective function value).
As an example, consider the SBM setup of Section \ref{sec:kb2}, with $\Lambda^{(1)}$ and $\Lambda^{(2)}$ defined as before, and $\Lambda^{(3)}$ set to be $\Lambda^{(2)}$.
If $m_1=m_2=m_3$, then the results of Section \ref{sec:kb2} imply that coarse matching would not recover the true permutation, while Theorem \ref{thm:secIV} implies clustered matching would recover the right permutation with high probability.

\begin{table}[b!]
\centering
\begin{tabular}{|c|c|c|c|c|c|c|}
\hline
$n$         & 50   & 50   & 50   & 100  & 100  & 100  \\ \hline
$m$         & 10   & 100  & 1000 & 10   & 100  & 1000 \\ \hline
$q=0.200$ & 1    & 1    & 1    & 1    & 1    & 1    \\ \hline
$q=0.225$ & 0.14 &1     & 1    &0.10  & 1    & 1    \\ \hline
$q=0.250$ & 0.40 & 0.50 & 0.42 & 0.12 & 1    & 0.19 \\ \hline
$q=0.275$ & 0.20 & 0.12 & 0.36 & 0.12 & 0.12 & 0.12 \\ \hline
$q=0.300$ & 0.20 & 0.30 & 0.12 & 0.07 & 0.09 & 0.07 \\ \hline
$q=0.325$ & 0.22 & 0.14 & 0.20 & 0.06 & 0.07 & 0.11 \\ \hline
$q=0.350$ & 0.14 & 0.24 & 0.14 & 0.08 & 0.08 & 0.08 \\ \hline
$q=0.375$ & 0.20 & 0.18 & 0.10 & 0.05 & 0.06 & 0.06 \\ \hline
$q=0.400$ & 0.10 & 0.14 & 0.10 & 0.05 & 0.07 & 0.10 \\ \hline
\end{tabular}
\caption{Table of matching accuracy in the single Erd\H os-R\'enyi background setting with $p=1/3$, averaged over 10 Monte Carlo iterates; similar results are obtained in the $p=0.5$ setting; see Appendix \ref{app:addexp} for detail.}
\label{tab:er13}
\vspace{-5mm}
\end{table}  


    \section{Simulations and Real Data Experiments}
    \label{sec:sim}
    \noindent
    We will now explore the impact of the three different strategies for matching $R$ to $C$ outlined in Section \ref{sec:problem_form}, 
    namely coarse matching, clustered matching, and fine matching.
    Note that in the experiments below, as computing the exact solution of the graph matching problem is often computationally intractable, we rely on the approximate graph matching algorithm, \texttt{SGM}, of \cite{ModFAQ}.
    This algorithm will use seeded vertices across $R$ and $C$ (those whose alignments via $P^*$ are a priori provided), as this will help us to hone in on when $f(P^*)$ is sub-optimal, which is our chief computational question.

\begin{figure}[b!]
    \centering
    \vspace{-5mm}
    \includegraphics[width = 0.45\textwidth]{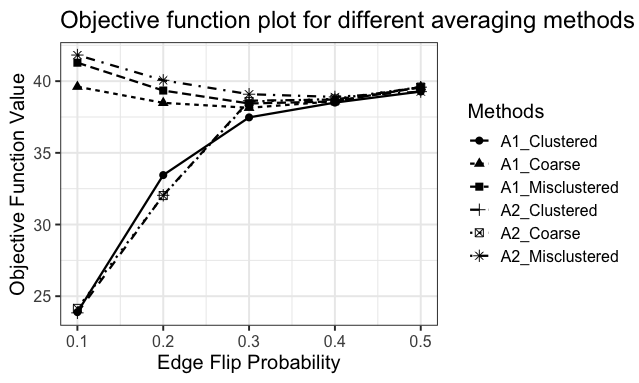}
    \caption{Objective function plot for different averaging methods for the two Erd\H os-R\'enyi background setting considered in Section \ref{sec:er}. 
    For each of the two  out-of-sample networks we perform coarse matching, clustered matching with its own cluster, and cluster matching with the incorrect cluster. 
    We plot the objective function of the match versus $q$ for each matching strategy/out-of-sample graph pair, averaged over 50 Monte Carlo iterates.
    Note that the $A_2$-clustered and $A_2$-coarse point values and subsequent lines are nearly identical, and are hard to distinguish; see Table \ref{tab:tabER2}.
    }
    \label{fig:figer2}
    \end{figure}
  
    \subsection{Matching in the ER model}
    \label{sec:er}
    \noindent
    We first consider the effectiveness of the coarse matching strategy in the $k=1$ setting in a simple Erd\H os–R\'enyi model with $n$ nodes and edge probability denoted by $p$. 
    In the $k=1$ setting, all in-sample networks are equally informative and averaging them into a background $C$ is sensible and recommended as long as the edge flipping probability is not too large.
    When the $Q$ matrix in Definition \ref{def:bitflip} is close to $1/2$, the in-sample and out-of-sample graphs become closer to independent, though this can be overcome to an extent by considering a large value of $m$.
    Formalizing this, we consider $B\sim\operatorname{ER}(n,p)$, and $A,S^{(1)}_i\stackrel{i.i.d.}{\sim}\operatorname{BF}(B,q)$, and we match $A$ (i.e., $P^*=I_n$) to $C$ using \texttt{SGM} with 5 randomly chosen seed vertices.
    In Figure \ref{fig:er2} we consider $p=1/3$ (similar results are obtained with $p=0.5$, see Appendix \ref{app:addexp} and Figure \ref{fig:er1} for detail), and we consider the effect of varying the number of nodes $n$ ($n=100$ in the figure, see Appendix \ref{app:addexp} for $n=50$ plots), and the number of in-sample graphs ($m=10$ in the left panel, $m=100$ in the middle panel, and $m=1000$ in the right panel).
    In each panel, we plot the \texttt{SGM} objective function value $f=\|A-PCP^T\|_F$ versus the value of the edge perturbation parameter $q$.
    When combined with the information in Table \ref{tab:er13} (see also Appendix \ref{app:addexp} for a full table), we see that for sufficiently small $q$ (here less than $0.2$), we will always recover the exact match, and the objective function is steadily increasing.
The jump in the objective function scores correspond to the point at which the \texttt{SGM} algorithm no longer recovers the true alignment, which is evidence for the true alignment no longer being optimal.
While subtle, we do see that this transition point occurs at a larger value of $q$ when $n$ and $m$ generally increase as expected.
The nature of the jump, and the relatively flat objective function value post-jump, across all the figures when \texttt{SGM} fails is indicative of the presence of phantom alignment strength after this critical threshold; see \cite{fishkind2021phantom} for further detail.

    We next consider the case of two backgrounds $B^{(1)}\sim\operatorname{ER}(n\!=\!80, p\!=\!0.2)$ and $B^{(2)}\sim\operatorname{ER}(n\!=\!80, p\!=\!0.4)$. 
    We let $S^{(1)}_1, \ldots, S^{(1)}_{m_1}$ i.i.d. sampled from $\operatorname{BF}(B^{(1)}, q)$ and $S^{(2)}_{1}, \ldots, S^{(2)}_{m_2}$ i.i.d. sampled from $\operatorname{BF}(B^{(2)},q)$ where $m_1=200$, $m_2=2000$ and $0<q<0.5$ is the edge flipping probability. 
    We draw two out-of-sample networks $A_i \sim \operatorname{BF}(B^{(i)}, q)$ for $i=1,2$ and match them with the full average of all the $S$'s, the average of just the $S^{(1)}$'s and the average of just $S^{(2)}$'s. 
    We plot the objective function of the match versus $q$ in Figure \ref{fig:figer2}, and provide the corresponding matching error rates (i.e., the proportion of labels incorrectly recovered) in Table \ref{tab:tabER2}; both are averaged over 50 Monte Carlo (MC) iterates.

We see that matching either graph $A_i$ to the coarse average, or the wrong cluster (i.e., matching $A_i$ to the average of $S^{(j)}$'s for $i\!\neq\! j$) yields poor matching accuracy and nearly uniformly high objective function value.
The exception is matching $A_2$ to the coarse mean when $q$ is small, due to the large proportion of type-2 graphs in the in-sample data, still enables a high fidelity matching.
As expected, matching to the correct in-sample cluster yields both better matching accuracy and better objective function value (compared to the wrong cluster matching), at least for modest values of $q$.
This points to the utility of using the class labels to locally average (or clustering) before matching, as the objective function value of matching to the class means can be used to identify the right class to match to which will then yield higher matching accuracy.

       \begin{table}[hb] 
    \begin{tabular}{|c|c|c|c|c|c|c|}
\hline
Method& $A_i$ class& q=0.1 & q=0.2 & q=0.3 & q=0.4 & q=0.5 \\ \hline
Coarse   &1& 0.080 & 0.076 &0.076 &0.076 &0.077 \\ \hline
Clustered&1&1.000 &0.737 &0.129 &0.084 &0.076 \\ \hline
Misclustered    &1&0.074 &0.073 &0.076 &0.075 &0.074 \\ \hline
Coarse  &2& 1.000 &1.000 &0.170 &0.093 &0.075 \\ \hline
Clustered     &2& 1.000 &0.987 &0.199 &0.089 &0.074 \\ \hline
Misclustered     &2& 0.074 &0.075 &0.075 &0.076 &0.074 \\ \hline
\end{tabular}
\caption{Table of matching accuracy in the two Erd\H os-R\'enyi background setting, averaged over 50 Monte Carlo iterates.
Values are rounded to three decimal places.
}
\label{tab:tabER2}
\vspace{-5mm}
\end{table}
    \subsection{Clustered matching in the COSIE model}
    \label{sec:cosiesim}
    \noindent
    Our theoretical results in the COSIE model show that when the score matrices are disordered in a similar direction, averaging across samples drawn from multiple backgrounds can produce inferior label recovery in the downstream out-of-sample matching task.
    If the score matrices are disordered in different enough directions, we expect that the noise in the score matrices could cancel (as in the SBM case of Section \ref{sec:kb2}), which would result in strong label recovery in the downstream out-of-sample matching task even when averaging a large number of wrong-cluster in-sample networks.

    \begin{figure}[t!]
    \centering
\includegraphics[width = 0.48\textwidth]{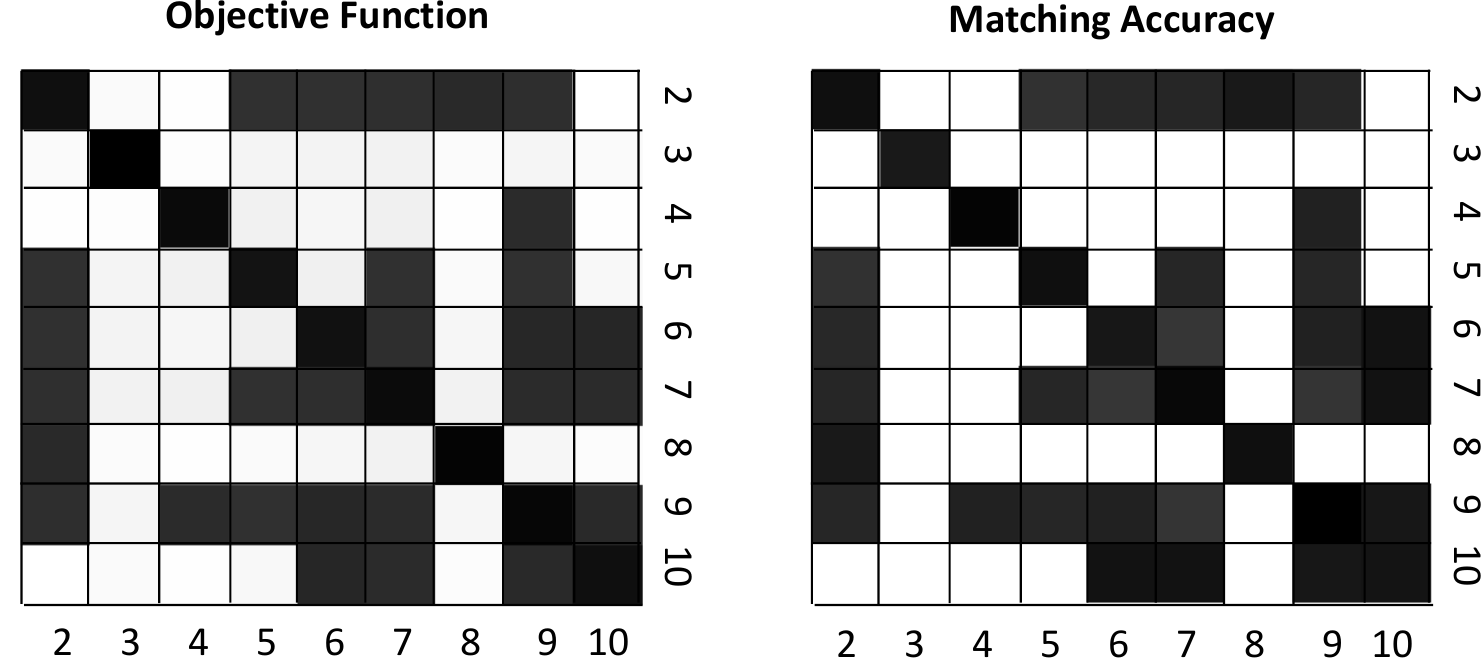}
        \caption{
        In the COSIE model considered in Section \ref{sec:cosiesim}, we consider matching $A$ to $C_{a,b}$ for $a\neq b$ ranging over $\{2,\ldots,10\}$.
            In the left (resp., right) heatmap, we plot the objective function value (resp., matching error rate)  obtained from \texttt{SGM} with 5 seeds.   In both heatmaps, lighter shade denotes smaller values/better matches while darker shade denotes larger values/worse matches; note that the diagonal blocks are not included as we assume $a\neq b$.}
        \label{fig:COSIE}
    \end{figure}
    
    We further explore this phenomenon in the following simple, yet illustrative experiment.
    We generate $k=10$ COSIE background graphs as follows:
    We consider $k=10$ independent $G_i\sim\operatorname{ER}(100,0.5)$ graphs (i.e., uniformly random graphs), and use the procedure in \cite{arroyo2021COSIE} to project these graphs into a common COSIE framework (i.e., finding a common $U$ and $R^{(i)}$'s such that $G_i\approx UR^{(i)}U^T$ and where each $R^{(i)}\in\mathbb{R}^{10\times 10}$).
    We then sample $B^{(i)}\sim\operatorname{COSIE}(U,R^{(i)})$, and 
    for each $i\in [10]$, we sample $m_i$ i.i.d. networks $S^{(i)}_1, \cdots, S^{(i)}_{m_{i}}$ from BF$(B^{(i)},0.1$). 
    We consider $A\sim\operatorname{BF}(B^{(1)},0.1$), and $m_1=10$, $m_i=5$ for $i\neq 1$.

    We then consider matching $A$ to $C_{a,b}$ where $C_{a,b}$ is formed via
    $C_{a,b}=\frac{1}{20}\sum_{i\in\{1,a,b\}}\sum_{j=1}^{m_i}S^{(i)}_j,$
    and where $a\neq b$ range over $\{2,\ldots,10\}$.
We plot a pair of heatmaps in Figure \ref{fig:COSIE} with indices representing values of $a, b$ chosen. 
In the left (resp., right) heatmap, we plot the objective function value (resp., matching error rate)  obtained from \texttt{SGM} with 5 seeds.  In both heatmaps, lighter shade denotes smaller values/better matches while darker shade denotes larger values/worse matches; note that the diagonal blocks are not included as we assume $a\neq b$.
    From the figure, we see a strong positive correlation between matching error rate and objective function score, and that which combination of background graphs are being averaged into $C_{a,b}$ is consequential and nuanced.  
    In Section \ref{sec:k2}, we saw that the nature of the backgrounds was crucial for determining whether a coarse matching would produce good results.
    In this example, similar to the SBM example considered in Section \ref{sec:kb2}, we consider $k=3$ and consider coarse matching of $(P^*)^TAP^*$ to $C$,  with the aim of better understanding when the coarse class averaging is beneficial/harmful for label recovery of the shuffled $A$.
    To this end, we set $m_1=m_2+m_3$, and we consider different combinations of background graphs $B^{(a)}$ and $B^{(b)}$ for representing classes 2 and 3 ($B^{(1)}$ will always represent class 1).
 
    As demonstrated in Theorem \ref{thm:cosie}, the wrong combination of in-sample backgrounds can lead to poor performance via coarse matching; this figure suggests that this phenomenon is neither uncommon nor straightforward.
    Indeed, while some background graph class pairs (e.g., (5,7)) have their order relative to $B^{(1)}$ combine to provide poor matching accuracy and large matching objective function, those same graphs paired differently (e.g., (5,6) and (7,8)) are relatively innocuous when averaged with the $S^{(1)}$'s, as the true alignment is still well-recovered even with coarse matching.
     
     \begin{figure*}[t!]
        \centering
        \includegraphics[width = 0.47\textwidth]{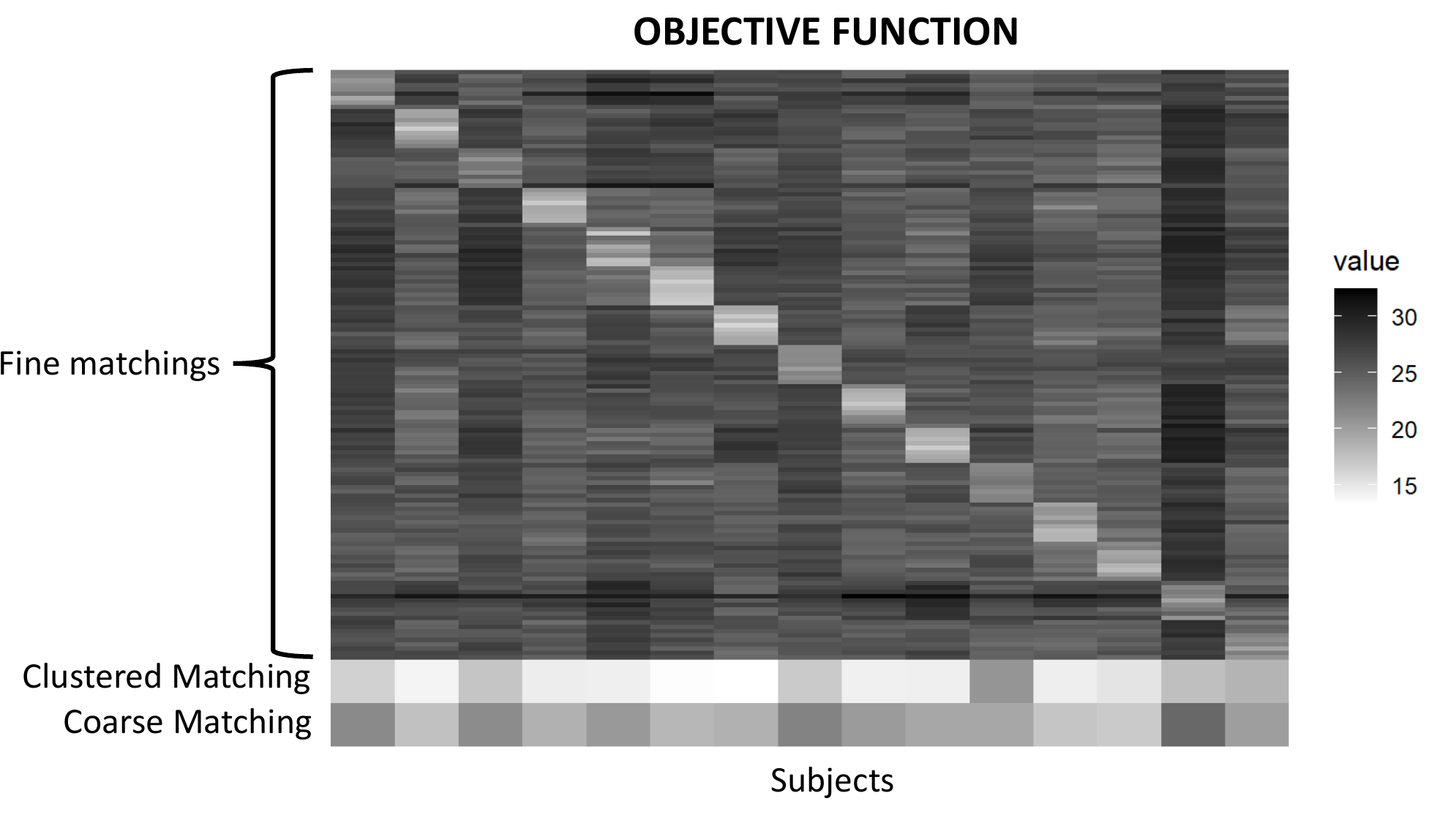}
        \includegraphics[width = 0.47\textwidth]{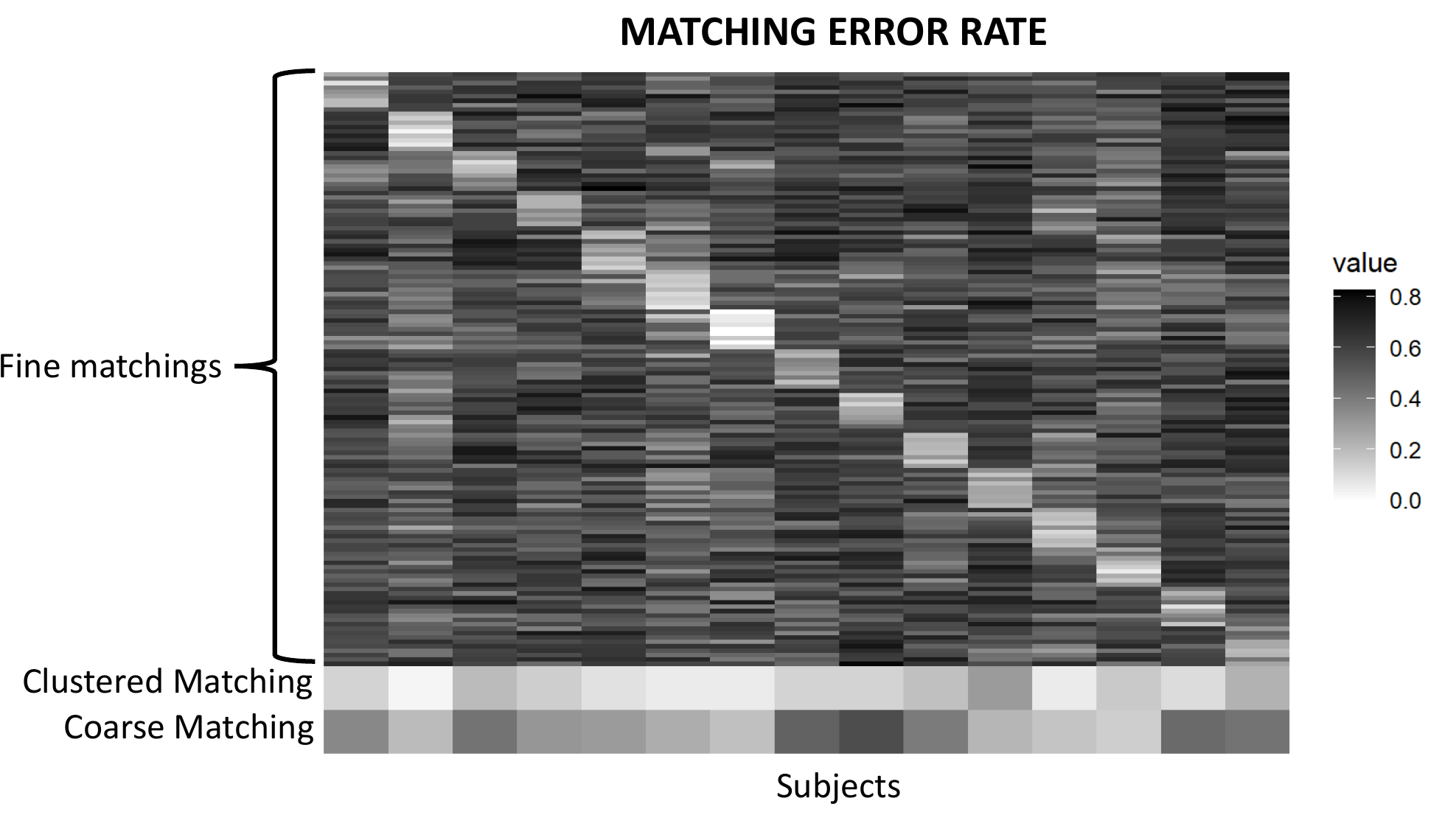}
        \caption{For each of the 15 out-of-sample brain networks, we match with: $(i)$ the average of all existing 135 graphs (coarse averaging); 
  $(ii)$ the average of the 9 in-sample graphs from the same subject (clustered averaging); 
  $(iii)$ each of the existing simulated graph (fine averaging). 
  In the top heatmap, we plot the objective function value obtained from \texttt{SGM} with 5 seeds, and in the bottom heatmap we plot the matching error rate.  In both heatmaps, lighter shade denotes smaller values/better matches while darker shade denotes larger values/worse matches.
    In each heatmap, the columns correspond to the 15 out-of-sample networks, with the rows corresponding to: top 135 (thinner) rows the fine matching with each in-sample network separately; the second-to-the bottom (thicker) row the clustered matching and the bottom (thicker) row the coarse matching result.}
        \label{fig:brains}
    \end{figure*}
    \subsection{Matching human connectomes}
    \label{sec:hnu1}
    \noindent
    We next consider a real data set of human connectomes from the HNU1 data repository \cite{zuo2014open}.
    In the dataset, for each of 30 subjects there are 10 test/retest DTMRI brain scans.
    The raw scans were processed via NeuroData's MRI Graphs (m2g) pipeline of \cite{kiar2018high} and registered to the Desikan atlas \cite{desikan2006automated}, yielding a 70 vertex weighted graph for each scan. 
    The graphs are a priori vertex-aligned both within and across subjects, with vertices in each graph representing regions of interest in the brain atlas, and with edges measuring the strength of the neuronal connections between regions. 
    The post-processed brain graphs are available from \url{neurodata.io}.
    
For our experiment, we randomly select 15 different subjects and their corresponding $15\times 10=150$ scans. 
We perform the experiment as follows: for each individual, we randomly take 9 brain graphs as the existing matched graphs (i.e., in-sample), with 1 brain graph assumed to be the out-of-sample network.
These 15 out-of-sample graphs will have both their class labels and vertex alignments (to the 135 in-sample graphs) treated as unknown/hidden in this experiment, with 
the goal then to recover the hidden class label (i.e., subject label) and vertex alignments for these out-of-sample graphs.
To recover the vertex alignments, for each of these 15 out-of-sample networks, we match them with: $(i)$ the average of all 135 in-sample graphs (coarse averaging); 
  $(ii)$ the average of the 9 in-sample graphs from the same subject (clustered averaging); and $(iii)$
  each of the in-sample graphs separately (fine averaging). 
  Note that while we used the true class/subject labels in our clustered averaging, these can be readily obtained via a simple k-means procedure applied to an embedded inter-graph distance matrix; see Appendix \ref{app:clust} for detail. 
  
  We plot heatmaps of the matching objective function and matching error in Figure \ref{fig:brains}.
    In the top heatmap, we plot the objective function value obtained from \texttt{SGM} with 5 seeds, and in the bottom heatmap we plot the matching error rate.  In both heatmaps, lighter shade denotes smaller values/better matches while darker shade denotes larger values/worse matches.
    In each heatmap, the columns correspond to the 15 out-of-sample networks, with the rows corresponding to: the fine matching (top 135 \textit{thinner} rows) with each in-sample network separately; the clustered matching (the second-to-the bottom \textit{thicker} row) and the coarse matching (the bottom row) results.
    From the figure, we see that for the majority of subjects, the clustered matching yields smaller objective function error and better matching accuracy than coarse matching (the subject in column 11 being the notable exception).  
    Moreover, we see that in some cases the best of the fine matchings yields better matching accuracy than even the clustered matching, though this is not always the case.
    For example, considering the matching accuracy at differing levels of granularity for a pair of subjects displayed in Table \ref{tab:brain}, we see
that for some patients the best fine matching yields the best matching accuracy while for others the clustered matching is best.
    \begin{table}[b!] 
    \begin{tabular}{|c|c|c|c|}
\hline
Subject & Coarse Matching& Clustered Matching& Fine Matching \\ \hline
0025435     & 0.8286& 0.9429     & 0.8857 \\ \hline
0025440     & 0.6000& 0.8143     & 0.8571\\ \hline
\end{tabular}
\caption{Matching accuracy for a pair of subjects across levels of granularity.}
\label{tab:brain}
\vspace{-5mm}
\end{table}
\begin{figure}[b!]
        \centering
        \vspace{-5mm}
        \includegraphics[width = 0.4\textwidth]{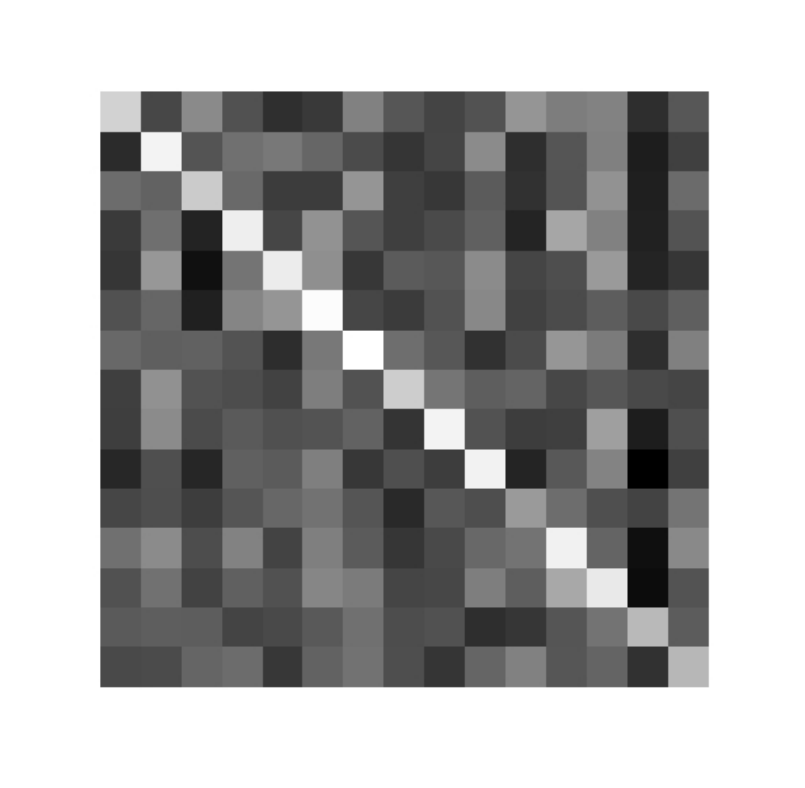}
        \vspace{-5mm}
        \caption{For each of the 15 out-of-sample brain networks, we
        plot a heatmap of the objective function obtained by \texttt{SGM} with 5 seeds by matching with each of the 15 in-sample cluster averages.
    In each heatmap, the columns correspond to the 15 out-of-sample networks, and the rows correspond to the 15 in-of-sample network averages (the diagonal corresponds to the matched indices).
    Larger values/worse matches in the heatmap are denoted by darker colors, with smaller values/better matches denoted by lighter colors.}
        \label{fig:brains2}
    \end{figure}
    
We next explore whether clustered averaging can be used to uncover the correct brain class labels as well.
This would be a key step for identifying the correct cluster to average to in Figure \ref{fig:brains2}.
To explore this, for each of the 15 out-of-sample brain networks, we
        plot a heatmap of the objective function obtained by \texttt{SGM} with 5 seeds by matching with each of the 15 in-sample cluster averages.
    In each heatmap, the columns correspond to the 15 out-of-sample networks, and the rows correspond to the 15 in-sample network averages (the diagonal corresponds to the matched indices).
    Larger values in the heatmap are denoted by darker colors.
    We indeed see that across the board, the cluster matching that obtains the best objective function is the one that matches the out-of-sample brains to the correct in-sample cluster average, pointing again to the validity of using this approach (with high fidelity clusters) for simultaneous classification and label alignment.
    While we do not suspect these brain graphs follow our posited bit-flipped model, the theory developed for our model nevertheless plays out in this real data setting:  the differences among the background connectome classes cause coarse matching to be less effective than clustered matching.
    This is as predicted by the theory, and clustered matching here provides both a computationally more efficient alternative to fine matching (that can produce better matching results) and an empirically better match than coarse matching.

    \section{Conclusion and discussion}
    \label{Sec:conc}
    \noindent
    We investigate strategies for recovering the vertex labels of an out-of-sample graph by using the information in a collection of vertex-aligned in-sample graphs.
    In both theory and synthetic/real data simulations, we explore the effectiveness of recovering the out-of-sample graph vertex labels by matching it to the in-sample collection at three levels of granularity.
    While it can be the case that the best method is to match the out-of-sample graph to all individual in-sample graphs and take the labels according to the matching result with smallest loss function, often this is too computationally expensive and the data-smoothing inherent to clustered matching often yields better alignment than the fine-grain matching. 
    At the other end of the granularity spectrum, in both theory and practice we demonstrate that  labeling the out-of-sample graph by matching it to the full average of all in-sample graphs can yield poor label recovery, especially in settings where there are significant differences in the structures across the in-sample graphs.

    Our proposed matching algorithm is a compromise between these two extremes.
    Our ``clustered matching" involves matching the out-of-sample graph individually to each class's average and labeling it via the matching result with smallest loss function. 
   A consequence of our theory is that given high enough fidelity classes, under mild model conditions the clustered matching will recover the right cluster and the right alignment with high probability.
   We also used both simulated as well as real world data to demonstrate the validity of the proposed algorithm as well as the advantage of clustered matching compared to the fine-grain and coarse-grain strategies outlined in Section \ref{sec:problem_form}.

    We also proposed the following possible extension and questions. 
    While we consider matching here to the usual sample average, there are a number of different notions of network means we could consider aligning to (e.g., Frechet means \cite{kolaczyk1,kolaczyk2020averages} or smoothed means \cite{tang2018connectome}).
    We next seek to relate this work to the phantom alignment strength conjecture proposed by Fishkind et. al. in \cite{fishkind2021phantom}. In particular, our result in the Erd\H os-R\'enyi model simulations (Fig. \ref{fig:er2}) showed matching objective functions similar to the ``hockey stick" matchability plots in \cite{fishkind2021phantom}. 
    Both our work and that in \cite{fishkind2021phantom} deal with edge-wise correlations, and we are working to unify our results and use our results and computations to support the foundation of the phantom alignment strength conjecture, ideally finding explanation or causation for the ``hockey sticks" matchability plots. 
    We would then be able to propose more precise conditions on when our three fore-mentioned matching strategies will behave similarly and when they will differ significantly. 

    Another important issue we want to explore is the edge-wise matchability of the out-of-sample graph. In particular, standing on a single edge level, it is hard to predict if the matching is exact for both clustered matching and fine matching. We want to find conditions or ways to verify if the edge-wise matching is indeed the exact one by looking at the edge mismatch level and finding computationally tractable remedies for misaligned structure. 
    Also, as in \cite{wu2021settling}, we want to explore the information theoretic recovery limitations of clustered versus coarse matching as well. 

Finally, it is important to note that if class labels are not known a priori, our proposed clustered matching relies heavily on a good graph clustering algorithm. 
If a clustering algorithm is provided, then our matching approaches are essentially standard GMP's and can be solved using existing methods and packages.
  
\bibliographystyle{plain}
\bibliography{biblio.bib}

\begin{thebibliography}{10}

\bibitem{Aldous2017SIR}
D.~Aldous.
\newblock The {SI} and {SIR} epidemics on general networks.
\newblock {\em Probab. Math. Stat.}, 37:229--234, 2017.

\bibitem{arroyo2021COSIE}
J.~Arroyo, A.~Athreya, J.~Cape, G.~Chen, Priebe~C. E., and J.~T. Vogelstein.
\newblock Inference for multiple heterogeneous networks with a common invariant
  subspace.
\newblock {\em J. Mach. Learn. Res.}, 22:1--49, 2021.

\bibitem{arroyo2021maximum}
J.~Arroyo, D.~L. Sussman, C.~E. Priebe, and V.~Lyzinski.
\newblock Maximum likelihood estimation and graph matching in errorfully
  observed networks.
\newblock {\em J. Comput. Graph. Stat.}, pages 1--13, 2021.

\bibitem{relion2019network}
J.~D. Arroyo~Reli{\'o}n, D.~Kessler, E.~Levina, and S.~F. Taylor.
\newblock Network classification with applications to brain connectomics.
\newblock {\em Ann. Appl. Stat.}, 13(3):1648, 2019.

\bibitem{Athreya2018RDPG}
A.~Athreya, D.~E. Fishkind, M.~Tang, C.~E. Priebe, Y.~Park, J.~T. Vogelstein,
  K.~Levin, V.~Lyzinski, Y.~Qin, and D.~Sussman.
\newblock Statistical inference on random dot product graphs: a survey.
\newblock {\em J. Mach. Learn. Res.}, 18:1--92, 2018.

\bibitem{athreya2013limit}
A.~Athreya, C.~E. Priebe, M.~Tang, V.~Lyzinski, D.~J. Marchette, and D.~L.
  Sussman.
\newblock A limit theorem for scaled eigenvectors of random dot product graphs.
\newblock {\em Sankhya A}, pages 1--18, 2013.

\bibitem{NetScienceBarabasi}
A.-L. Barab{\'a}si.
\newblock {\em Network Science}.
\newblock Cambridge University Press, 2016.

\bibitem{barak2019nearly}
B.~Barak, C.~Chou, Z.~Lei, T.~Schramm, and Y.~Sheng.
\newblock (nearly) efficient algorithms for the graph matching problem on
  correlated random graphs.
\newblock {\em Adv. in Neural Infor. Proc. Sys.}, 32:9190--9198, 2019.

\bibitem{Bickel2009}
P.~J. Bickel and A.~Chen.
\newblock {A nonparametric view of network models and Newman-Girvan and other
  modularities.}
\newblock {\em Proc. Natl. Acad. Sci. U.S.A.}, 106:21068--73, 2009.

\bibitem{bickel2011method}
P.~J. Bickel, A.~Chen, and E.~Levina.
\newblock The method of moments and degree distributions for network models.
\newblock {\em Ann. Stat.}, 39(5):2280--2301, 2011.

\bibitem{Borgwardt2005KernalClass}
K.~M. Borgwardt, C.~S. Ong, S.~Schönauer, S.~V. Vishwanathan, A.~J. Smola, and
  H.~P. Kriegel.
\newblock Protein function prediction via graph kernels.
\newblock {\em Bioinformatics}, 2005.

\bibitem{Broder2000WWW}
A.~Broder, R~Kumar, F~Maghoul, P~Raghavan, S~Rajagopalan, R~Stata, A~Tomkins,
  and J~Wiener.
\newblock Graph structure in the web.
\newblock {\em The Structure and Dynamics of Networks}, pages 309--320, 2000.

\bibitem{sporns_complex}
E.~Bullmore and O.~Sporns.
\newblock Complex brain networks: Graph theoretical analysis of structural and
  functional systems.
\newblock {\em Nature Rev. Neurosci}, 10:186--198, 2009.

\bibitem{chung2021statistical}
J.~Chung, E.~Bridgeford, J.~Arroyo, B.~D. Pedigo, A.~Saad-Eldin,
  V.~Gopalakrishnan, L.~Xiang, C.~E. Priebe, and J.~T. Vogelstein.
\newblock Statistical connectomics.
\newblock {\em Annu. Rev. Stat. Appl.}, 8:463--492, 2021.

\bibitem{ConFogSanVen2004}
D.~Conte, P.~Foggia, C.~Sansone, and M.~Vento.
\newblock Thirty years of graph matching in pattern recognition.
\newblock {\em Int. J. Pattern Recognit. Artif. Intell.}, 18(03):265--298,
  2004.

\bibitem{cullina2016improved}
D.~Cullina and N.~Kiyavash.
\newblock Improved achievability and converse bounds for {E}rdos-{R}enyi graph
  matching.
\newblock In {\em ACM SIGMETRICS Performance Evaluation Review}, volume 44 (1),
  pages 63--72. ACM, 2016.

\bibitem{cullina2017exact}
D.~Cullina and N.~Kiyavash.
\newblock Exact alignment recovery for correlated {E}rdos {R}enyi graphs.
\newblock {\em arXiv preprint arXiv:1711.06783}, 2017.

\bibitem{desikan2006automated}
R.~S. Desikan, F.~S{\'e}gonne, B.~Fischl, B.~T. Quinn, B.~C. Dickerson,
  D.~Blacker, R.~L. Buckner, A.~M. Dale, R.~P. Maguire, B.~T. Hyman, et~al.
\newblock An automated labeling system for subdividing the human cerebral
  cortex on mri scans into gyral based regions of interest.
\newblock {\em Neuroimage}, 31(3):968--980, 2006.

\bibitem{draves2020bias}
B.~Draves and D.~L. Sussman.
\newblock Bias-variance tradeoffs in joint spectral embeddings.
\newblock {\em arXiv preprint arXiv:2005.02511}, 2020.

\bibitem{Duvenaud2015CNNClass}
D.~K. Duvenaud, D.~Maclaurin, J.~Iparraguirre, R.~Bombarell, T.~Hirzel,
  A.~Aspuru-Guzik, and R.~P. Adams.
\newblock Convolutional networks on graphs for learning molecular fingerprints.
\newblock In {\em Adv. in Neural Infor. Proc. Sys.}, volume~28. Curran
  Associates, Inc., 2015.

\bibitem{ErdRen1963}
P.~Erd\H{o}s and A.~R\'{e}nyi.
\newblock Asymmetric graphs.
\newblock {\em Acta Mathematica Academiae Scientiarum Hungarica},
  14(3--4):295--315, 1963.

\bibitem{fan2020spectral}
Z.~Fan, C.~Mao, Y.~Wu, and J.~Xu.
\newblock Spectral graph matching and regularized quadratic relaxations:
  Algorithm and theory.
\newblock In {\em Inter. Conf. on Mach. Learn.}, pages 2985--2995. PMLR, 2020.

\bibitem{fishkind2021phantom}
D.~E. Fishkind, F.~Parker, H.~Sawczuk, L.~Meng, E.~Bridgeford, A.~Athreya,
  C.~Priebe, and V.~Lyzinski.
\newblock The phantom alignment strength conjecture: practical use of graph
  matching alignment strength to indicate a meaningful graph match.
\newblock {\em Appl. Netw. Sci.}, 6(1):1--27, 2021.

\bibitem{ModFAQ}
D.E. Fishkind, S.~Adali, H.G. Patsolic, L.~Meng, D.~Singh, V.~Lyzinski, and
  C.E. Priebe.
\newblock Seeded graph matching.
\newblock {\em Pattern Recognit.}, 87:203 -- 215, 2019.

\bibitem{FogPerVen2014}
P.~Foggia, G.~Percannella, and M.~Vento.
\newblock Graph matching and learning in pattern recognition in the last 10
  years.
\newblock {\em Int. J. Pattern Recognit. Artif. Intell.}, 28(01):1450001, 2014.

\bibitem{kolaczyk1}
C.~E. Ginestet, J.~Li, P.~Balachandran, S.~Rosenberg, and E.~D. Kolaczyk.
\newblock Hypothesis testing for network data in functional neuroimaging.
\newblock {\em Ann. Appl. Stat.}, pages 725--750, 2017.

\bibitem{goldenberg2010survey}
A.~Goldenberg, A.~X. Zheng, S.~E. Fienberg, and E.~M. Airoldi.
\newblock A survey of statistical network models.
\newblock {\em Found. Trends Mach. Learn.}, 2(2):129--233, 2010.

\bibitem{wrg2}
W.~R. Gray and et~al.
\newblock Migraine: Mri graph reliability analysis and inference for
  connectomics.
\newblock {\em GlobalSIP}, 2013.

\bibitem{Hoff2002}
P.~D. Hoff, A.~E. Raftery, and M.~S. Handcock.
\newblock {Latent space approaches to social network analysis}.
\newblock {\em J. Am. Stat. Assoc.}, 97(460):1090--1098, 2002.

\bibitem{hoffman2003variation}
A.~J. Hoffman and H.~W. Wielandt.
\newblock The variation of the spectrum of a normal matrix.
\newblock In {\em Selected Papers Of Alan J Hoffman: With Commentary}, pages
  118--120. World Scientific, 2003.

\bibitem{Holland1983}
P.~W. Holland, K.~Laskey, and S.~Leinhardt.
\newblock {Stochastic blockmodels: First steps}.
\newblock {\em Social Networks}, 5(2):109--137, 1983.

\bibitem{kiar2018high}
G.~Kiar, E.~W. Bridgeford, W.~R.~G. Roncal, V.~Chandrashekhar, D.~Mhembere,
  S.~Ryman, X.~Zuo, D.~S. Margulies, R.~C. Craddock, C.~E. Priebe, R.~Jung,
  V.~Calhoun, B.~Caffo, R.~Burns, M.~P. Milham, and J.~Vogelstein.
\newblock A high-throughput pipeline identifies robust connectomes but
  troublesome variability.
\newblock {\em bioRxiv}, page 188706, 2018.

\bibitem{kolaczyk2014statistical}
E.~D. Kolaczyk and G.~Cs{\'a}rdi.
\newblock {\em Statistical analysis of network data with R}, volume~65.
\newblock Springer, 2014.

\bibitem{kolaczyk2020averages}
E.~D. Kolaczyk, L.~Lin, S.~Rosenberg, J.~Walters, and J.~Xu.
\newblock Averages of unlabeled networks: Geometric characterization and
  asymptotic behavior.
\newblock {\em Ann. Stat.}, 48(1):514--538, 2020.

\bibitem{kolda2009tensor}
T.~G. Kolda and B.~W. Bader.
\newblock Tensor decompositions and applications.
\newblock {\em SIAM review}, 51(3):455--500, 2009.

\bibitem{levin2017central}
K.~Levin, A.~Athreya, V.~Tang, M.and~Lyzinski, Y.~Park, and C.~E. Priebe.
\newblock A central limit theorem for an omnibus embedding of multiple random
  graphs and implications for multiscale network inference.
\newblock {\em arXiv preprint arXiv:1705.09355}, 2017.

\bibitem{lyzinski2016information}
V.~Lyzinski.
\newblock Information recovery in shuffled graphs via graph matching.
\newblock {\em IEEE Trans. Inf. Theory}, 64(5):3254--3273, 2018.

\bibitem{lyzinski2014seeded}
V.~Lyzinski, S.~Adali, J.~T. Vogelstein, Y.~Park, and C.~E. Priebe.
\newblock Seeded graph matching via joint optimization of fidelity and
  commensurability.
\newblock {\em arXiv preprint arXiv:1401.3813}, 2014.

\bibitem{lyz2016relax}
V.~Lyzinski, D.~E. Fishkind, M.~Fiori, J.~T. Vogelstein, C.~E. Priebe, and
  G.~Sapiro.
\newblock Relax at your own risk.
\newblock {\em IEEE Trans. Pattern Anal. Mach. Intell.}, pages 60--73, 2016.

\bibitem{lyzinski2020matchability}
V.~Lyzinski and D.~L. Sussman.
\newblock Matchability of heterogeneous networks pairs.
\newblock {\em Information and Inference: A Journ. of the IMA}, 9(4):749--783,
  2020.

\bibitem{Mishra2014Social}
S.~Mishra, R.~Borboruah, B.~Choudhury, and S.~Rakshit.
\newblock Modeling of social network using graph theoretical approach.
\newblock {\em International Journ. of Computer Applications}, pages 34--37,
  2014.

\bibitem{Newman2002Disease}
M.~E.~J. Newman.
\newblock Spread of epidemic disease on networks.
\newblock {\em Phys. Rev. E}, 66:016128, 2002.

\bibitem{Newman2003ComplexNetwork}
M.~E.~J. Newman.
\newblock The structure and function of complex networks.
\newblock {\em SIAM REVIEW}, 45(2):167--256, 2003.

\bibitem{newman2018networks}
M.~E.~J. Newman.
\newblock {\em Networks}.
\newblock Oxford university press, 2018.

\bibitem{nielsen2018multiple}
A.~M. Nielsen and D.~Witten.
\newblock The multiple random dot product graph model.
\newblock {\em arXiv preprint arXiv:1811.12172}, 2018.

\bibitem{NiepertCNNClass}
M.~Niepert, M.~Ahmed, and K.~Kutzkov.
\newblock Learning convolutional neural networks for graphs.
\newblock In {\em Proc. 33rd Inter. Conf. on Mach. Learn.}, volume~48 of {\em
  Proc. Mach. Learn. Res.}, pages 2014--2023. PMLR, 20--22 Jun 2016.

\bibitem{Nikolentzos2017MatchingNE}
G.~Nikolentzos, P.~Meladianos, and M.~Vazirgiannis.
\newblock Matching node embeddings for graph similarity.
\newblock In {\em AAAI}, 2017.

\bibitem{onaran2016optimal}
E.~Onaran, S.~Garg, and E.~Erkip.
\newblock Optimal de-anonymization in random graphs with community structure.
\newblock In {\em 2016 50th Asilomar Conf. on Signals, Systems and Computers},
  pages 709--713. IEEE, 2016.

\bibitem{Gross2011OPAN}
P.~Pedarsani and M.~Grossglauser.
\newblock On the privacy of anonymized networks.
\newblock In {\em Proc. of the 17th ACM SIGKDD}, pages 1235--1243, 2011.

\bibitem{priebe2015statistical}
C.~E. Priebe, D.~L. Sussman, M.~Tang, and J.~T. Vogelstein.
\newblock Statistical inference on errorfully observed graphs.
\newblock {\em J. Comput. Graph. Stat.}, 24(4):930--953, 2015.

\bibitem{racz2021correlated}
M.~Racz and A.~Sridhar.
\newblock Correlated stochastic block models: Exact graph matching with
  applications to recovering communities.
\newblock {\em Adv. in Neural Infor. Proc. Sys.}, 34, 2021.

\bibitem{rand1971objective}
W.~M. Rand.
\newblock Objective criteria for the evaluation of clustering methods.
\newblock {\em J. Am. Stat. Assoc.}, 66(336):846--850, 1971.

\bibitem{ross2011fundamentals}
N.~Ross.
\newblock Fundamentals of {S}tein’s method.
\newblock {\em Probab. Surv.}, 8:210--293, 2011.

\bibitem{SABARISH2020Hierarchical}
B.~A. Sabarish, R.~Karthi, and K.~T. Gireesh.
\newblock Graph similarity-based hierarchical clustering of trajectory data.
\newblock {\em Procedia Computer Science}, 171:32--41, 2020.
\newblock Third Intern. Conf. Comp. Network Comm. (CoCoNet'19).

\bibitem{seshadhri2020impossibility}
C~Seshadhri, A.~Sharma, A.~Stolman, and A.~Goel.
\newblock The impossibility of low-rank representations for triangle-rich
  complex networks.
\newblock {\em Proc. Natl. Acad. Sci. U.S.A.}, 117(11):5631--5637, 2020.

\bibitem{shervashidze2011WLkernel}
N.~Shervashidze, P.~Schweitzer, E.J.V. Leeuwen, K.~Mehlhorn, and K.~M.
  Borgwardt.
\newblock Weisfeiler-lehman graph kernels.
\newblock {\em J. Mach. Learn. Res.}, 12(77):2539--2561, 2011.

\bibitem{sussman2018matched}
D.~L. Sussman, Y.~Park, C.~E. Priebe, and V.~Lyzinski.
\newblock Matched filters for noisy induced subgraph detection.
\newblock {\em IEEE Trans. Pattern Anal. Mach. Intell.}, 42(11):2887--2900,
  2019.

\bibitem{tang14:_semipar}
M.~Tang, A.~Athreya, D.~L. Sussman, V.~Lyzinski, and C.~E. Priebe.
\newblock A semiparametric two-sample hypothesis testing for random dot product
  graphs.
\newblock arXiv preprint. \url{http://arxiv.org/abs/1403.7249}, 2014.

\bibitem{tang2017asymptotically}
M.~Tang, J.~Cape, and C.~E. Priebe.
\newblock Asymptotically efficient estimators for stochastic blockmodels: The
  naive mle, the rank-constrained mle, and the spectral.
\newblock {\em Bernoulli}, 28(2):1049--1073, 2022.

\bibitem{tang2018limit}
M.~Tang and C.~E. Priebe.
\newblock Limit theorems for eigenvectors of the normalized laplacian for
  random graphs.
\newblock {\em Ann. Stat.}, 46(5):2360--2415, 2018.

\bibitem{tang2018connectome}
R.~Tang, M.~Ketcha, A.~Badea, E.~D. Calabrese, D.~S. Margulies, J.~T.
  Vogelstein, C.~E. Priebe, and D.~L. Sussman.
\newblock Connectome smoothing via low-rank approximations.
\newblock {\em IEEE Trans. Med. Imaging}, 38(6):1446--1456, 2018.

\bibitem{vaca2022systematic}
F.~Vaca-Ram{\'\i}rez and T.~P. Peixoto.
\newblock Systematic assessment of the quality of fit of the stochastic block
  model for empirical networks.
\newblock {\em Physical Review E}, 105(5):054311, 2022.

\bibitem{vogelstein2011shuffled}
J.~T. Vogelstein and C.~E. Priebe.
\newblock Shuffled graph classification: Theory and connectome applications.
\newblock {\em Journ. of Classification}, 32(1):3--20, 2015.

\bibitem{vogelstein2013graph}
J.~T. Vogelstein, W.~G. Roncal, R.~J. Vogelstein, and C.~E. Priebe.
\newblock Graph classification using signal-subgraphs: Applications in
  statistical connectomics.
\newblock {\em IEEE Trans. Pattern Anal. Mach. Intell.}, 35(7):1539--1551,
  2013.

\bibitem{wang2021learning}
Lu~Wang, Feng~Vankee Lin, Martin Cole, and Zhengwu Zhang.
\newblock Learning clique subgraphs in structural brain network classification
  with application to crystallized cognition.
\newblock {\em Neuroimage}, 225:117493, 2021.

\bibitem{wang2019joint}
S.~Wang, J.~Arroyo, J.~T. Vogelstein, and C.~E. Priebe.
\newblock Joint embedding of graphs.
\newblock {\em IEEE Trans. Pattern Anal. Mach. Intell.}, 43(4):1324--1336,
  2019.

\bibitem{wolfe_olhede_graphon}
P.~J. Wolfe and S.~C. Olhede.
\newblock Nonparametric graphon estimation.
\newblock ArXiv preprint at \url{http://arxiv.org/abs/1309.5936}, 2013.

\bibitem{wu2021settling}
Y.~Wu, J.~Xu, and S.~H. Yu.
\newblock Settling the sharp reconstruction thresholds of random graph
  matching.
\newblock {\em IEEE Trans. Inf. Theory}, 2022.

\bibitem{yan2016short}
J.~Yan, X.~Yin, W.~Lin, C.~Deng, H.~Zha, and X.~Yang.
\newblock A short survey of recent advances in graph matching.
\newblock In {\em 2016 Proc. Int. Conf. Multimed. Inf. Retr.}, pages 167--174.
  ACM, 2016.

\bibitem{zhou2021dynamic}
Y.~Zhou and H.-G. M{\"u}ller.
\newblock Dynamic network regression.
\newblock {\em arXiv preprint arXiv:2109.02981}, 2021.

\bibitem{zuo2014open}
X.~Zuo, J.~S. Anderson, P.~Bellec, R.~M. Birn, B.~B. Biswal, J.~Blautzik,
  J.~Breitner, R.~L. Buckner, V.~D. Calhoun, F.~X. Castellanos, et~al.
\newblock An open science resource for establishing reliability and
  reproducibility in functional connectomics.
\newblock {\em Scientific data}, 1(1):1--13, 2014.

\end{thebibliography}

\vspace{-8mm}
\begin{IEEEbiography}[{\includegraphics[width=1in,keepaspectratio]{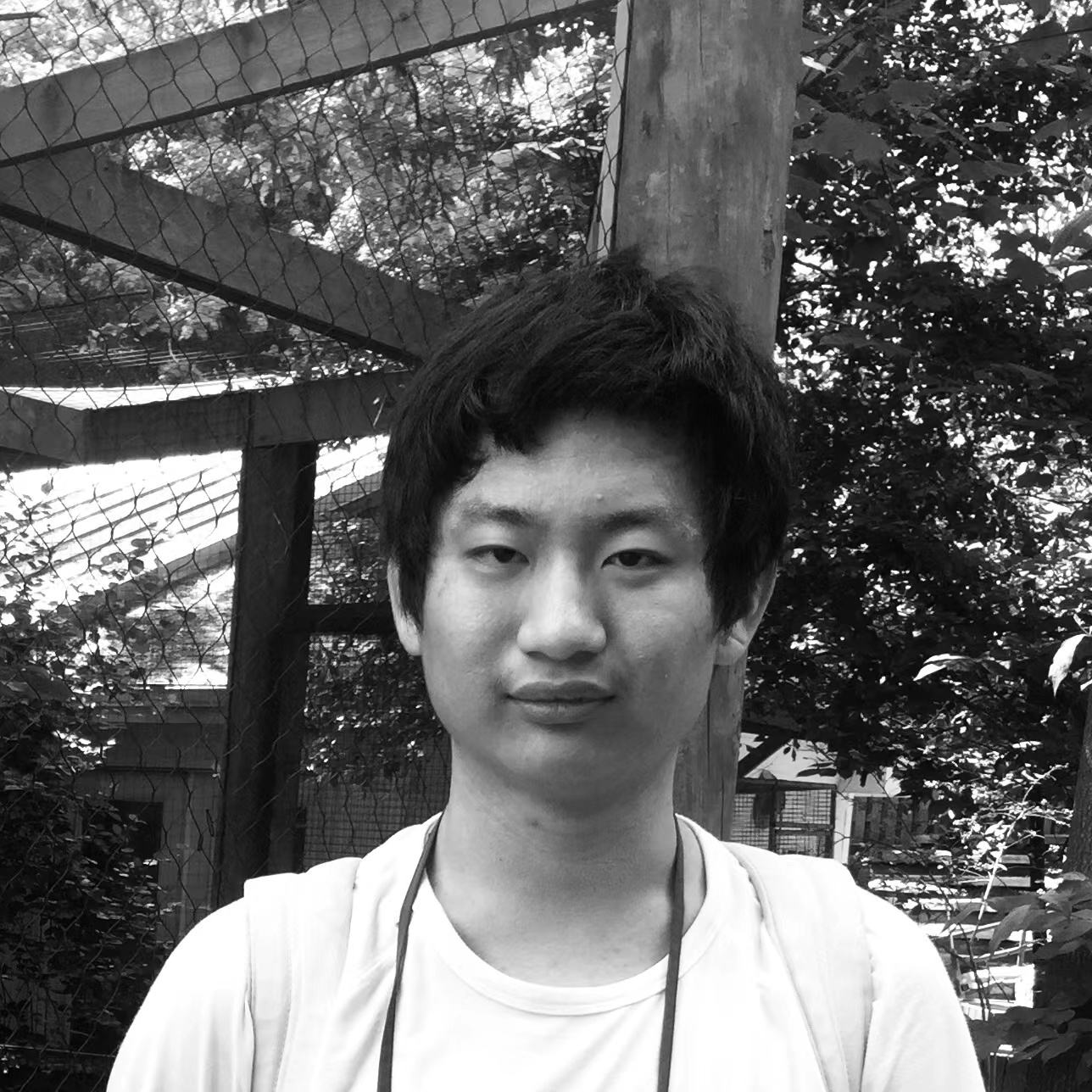}}]{Zhirui Li}
 received the BS degrees in mathematics and statistics from the University of Iowa, in
2020. He is now a doctoral student with the AMSC program at the University of Maryland, College Park. His advisor is Dr. Vince Lizinski, Associate Professor in the Department of Mathematics. His research interests include graph matching, stochastic processes, and machine learning.
\end{IEEEbiography}

\vspace{-8mm}
\begin{IEEEbiography}[{\includegraphics[width=1in,clip,keepaspectratio]{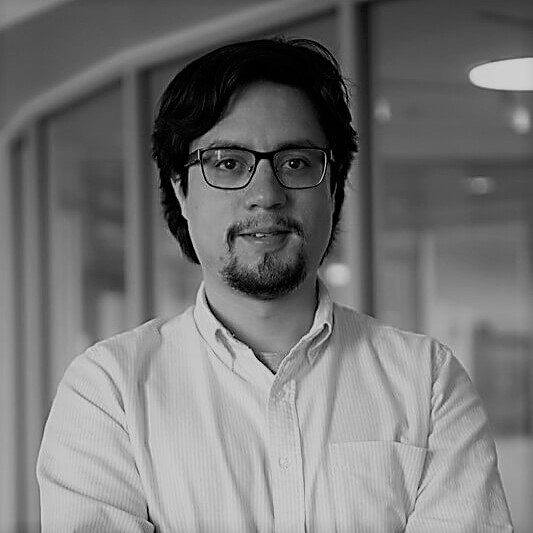}}]{Jes\'us Arroyo}  received the BS degrees in applied
mathematics and computer engineering
from the Instituto Tecnol\'ogico Aut \'onomo de
M\'exico (ITAM) in 2013, and the MA and PhD
degrees from the Department of Statistics at the
University of Michigan, Ann Arbor, in 2018. After
that, he was a postdoctoral fellow in the Center
for Imaging Science at Johns Hopkins University
from 2018 to 2020, and in the Department of
Mathematics at the University of Maryland, College
Park from 2020 to 2021. He is currently an assistant professor in the Department of Statistics at Texas A\&M
University since 2021. His research interests include include statistical network
analysis, machine learning, high-dimensional data analysis, and
applications to neuroimaging.
\end{IEEEbiography}

\vspace{-8mm}

\begin{IEEEbiography}[{\includegraphics[width=1in,height=1.25in,clip,keepaspectratio]{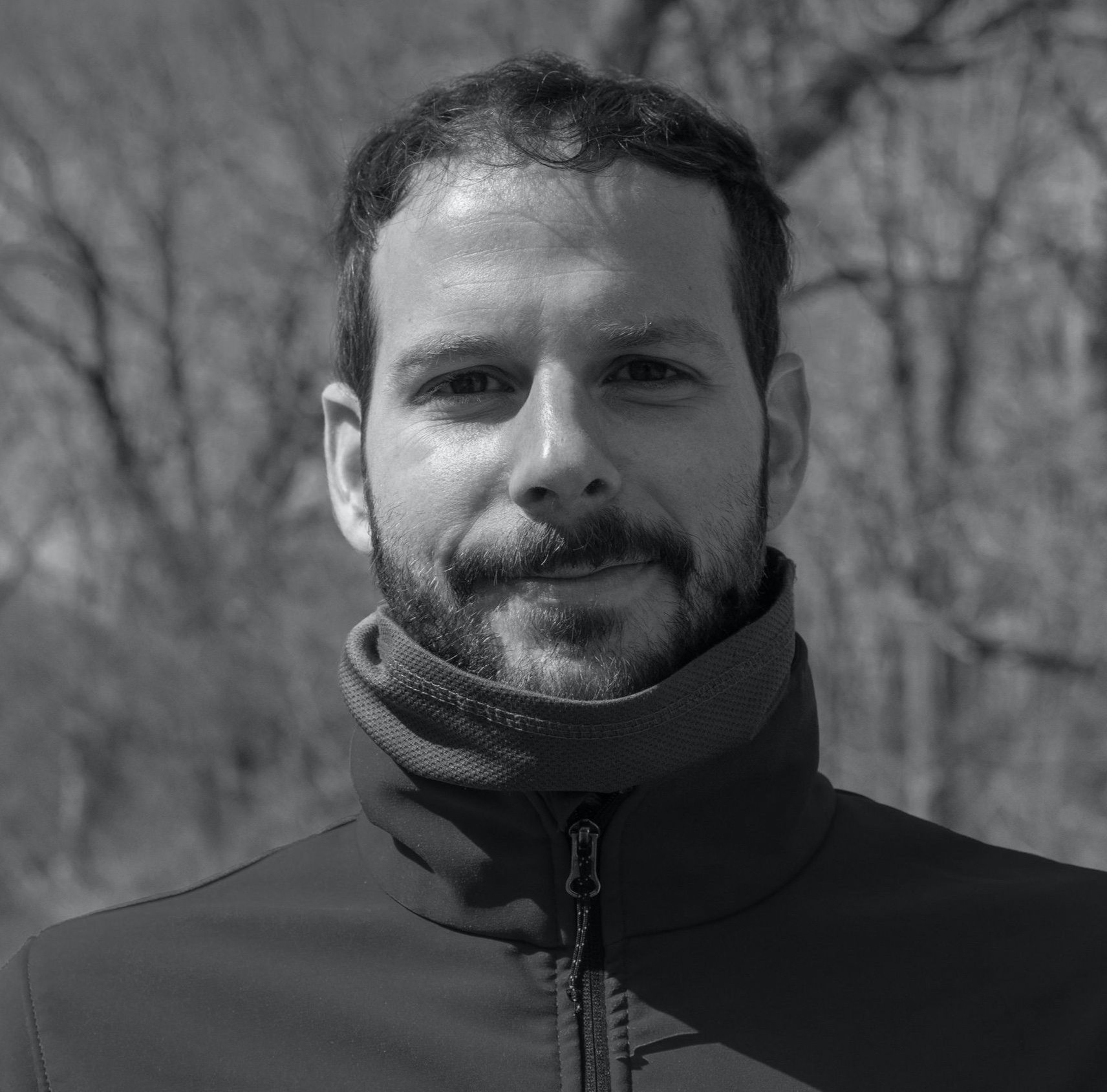}}]{Konstantinos Pantazis} received the BSc degree in Mathematics, from the National and Kapodistrian University of Athens, in
2016, and the Ph.D degree in Mathematics and Statistics from the University of Maryland, College Park in May 2022. During 2022-2023, he is a postdoctoral fellow with the Department of Applied Mathematics and Statistics
(AMS) at Johns Hopkins University.
His research areas of interest include multiscale statistical network inference, multiple graph matching and time series of networks.
Webpage: \url{https://kpantazis.github.io/}.
\end{IEEEbiography}

\vspace{-8mm}
\begin{IEEEbiography}[{\includegraphics[width=1in,height=1.25in,clip,keepaspectratio]{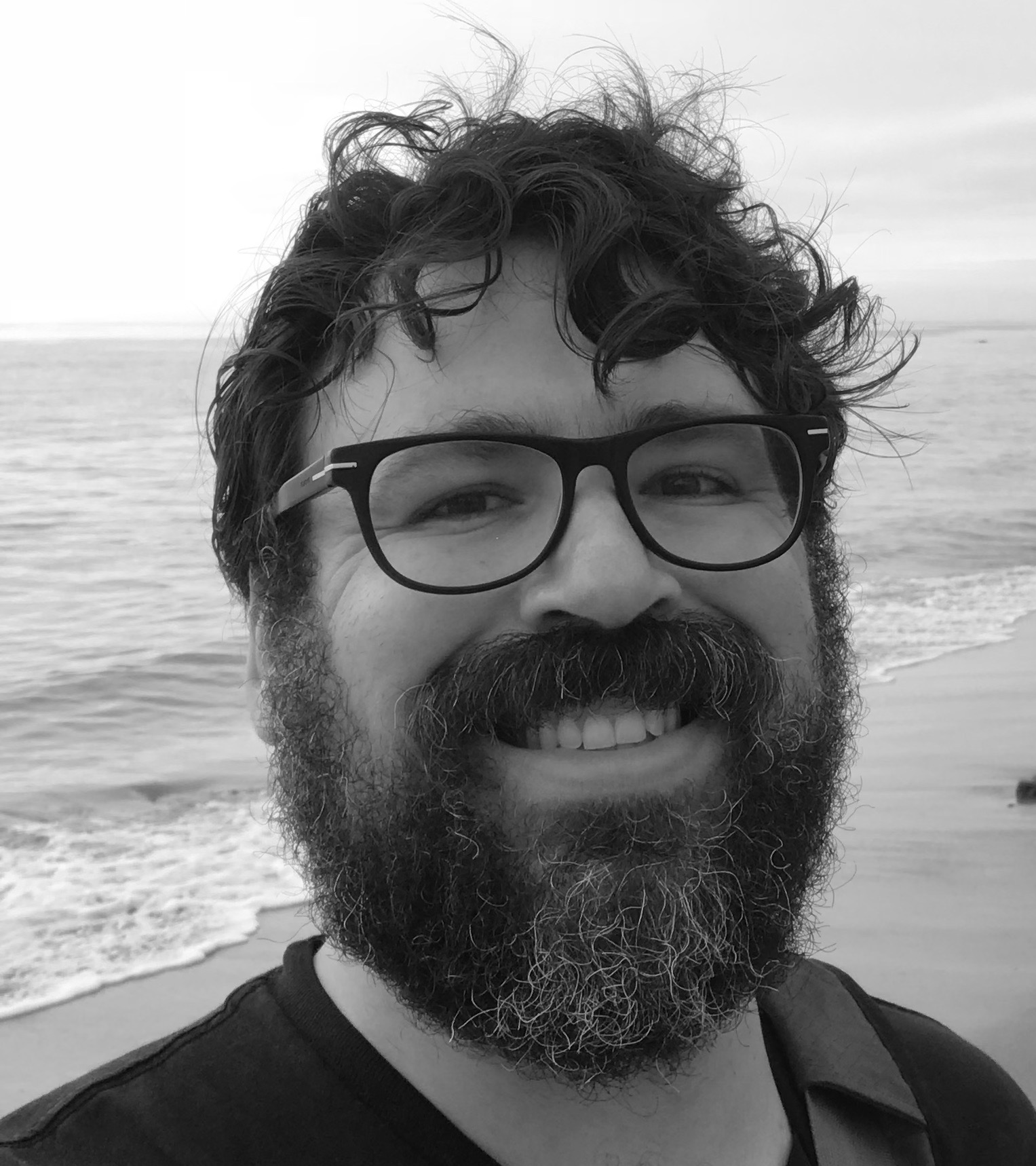}}]{Vince Lyzinski}
 received the BSc degree in mathematics from the University of Notre Dame, in
2006, the MA degree in mathematics from Johns
Hopkins University (JHU), in 2007, the MSE
degree in applied mathematics and statistics
from JHU, in 2010, and the PhD degree in applied
mathematics and statistics from JHU, in 2013.
From 2013-2014 he was a postdoctoral fellow
with the Applied Mathematics and Statistics
(AMS) Department, JHU. During 2014-2017, he
was a senior research scientist with the JHU
HLTCOE and an assistant research professor with the AMS Department, JHU.
From 2017-2019, he was on the Faculty in the Department of Mathematics
and Statistics at the University of Massachusetts Amherst. Since 2019 he has been on the Faculty in the Department of Mathematics at the University of Maryland, College Park, where he is currently an Associate Professor. His research interests include graph matching, statistical
inference on random graphs, pattern recognition, dimensionality reduction, stochastic processes, and high-dimensional data analysis.
\end{IEEEbiography}

\onecolumn
\section{Appendix}

\subsection{Notation used throughout the appendix}
Throughout the appendix, we will observe that 
\begin{align}
f(P)\!-\!f(P^*)\!
=\sum_{\substack{h,\ell,\,\text{ s.t.} \\\{\sigma(h),\sigma(\ell)\}\neq\{h,\ell\}}}\left(\sum_{i,j}S_i^{(j)}[h,\ell]\right)(A[\sigma(h),\sigma(\ell) ]\!-\!A[h,\ell]),
\label{eq:fdiff}\end{align}
where $\sigma$ is the permutation associated with $P^\bigstar=P(P^*)^T$.
This expansion is essential for applying the concentration inequalities appearing throughout the manuscript.

\subsection{Connection between Definition \ref{def:bitflip} and Erd\H os-R\'enyi matchability}
\label{app:ER}
The model in Definition~\ref{def:bitflip} can be used to study the phenomenon of graph matchability/graph de-anonymization.
Loosely stated, if graphs $A$ and $B$ in $\mathcal{G}_n$ have true, but latent, alignment $P^*$, graph matchability is concerned with understanding the conditions (often in terms of the edge correlation across networks) under which 
$$\{P^*\}=
\text{argmin}_{P\in\Pi_n}\|A-PBP^T\|_F.$$
(determining whether there is enough signal to match the vertices of two random graphs).
Considering $B\sim \mathrm{ER}(n,p)$ in Definition \ref{def:bitflip} and $Q=sJ_n$ (where $J_n$ is the hollow $n\times n$ matrix with all off-diagonal entries identically equal to 1),  matchability in the classical correlated Erd\H os-R\'enyi model is obtained by considering alignments of $S_1$ and (a shuffled) $S_2$, where $S_1,S_2|A\stackrel{i.i.d.}{\sim}\operatorname{BF}(B,Q)$.
Indeed, (no longer conditioning on $A$) in this case $S_1$ and $S_2$ would both have ER$(n,p(1-s)+s(1-p))$ distributions and the edge-wise correlation is given by 
$$\text{corr}(S_{1,ij},S_{2,ij})=\frac{p(1-p)(1-2s)^2 }{(p+s-2sp)(1-p-s+2sp) }.$$
Here, sharp matchability thresholds are established in \cite{wu2021settling, cullina2017exact} in terms of $s$ and $p$ (i.e., in terms of the correlation across networks).


\subsection{Proof of Theorem 1:}
\label{app:pf1}
\begin{proof}
We will use Stein's method to prove this result; principally Theorem 3.6 in \cite{ross2011fundamentals}.
We say that a collection of random variables $(X_1,\cdots,X_n)$ has \emph{dependency neighborhoods} $N_i\subset[n]$ for $i\in[n]$ if for each $i$, $X_i$ is independent of $\{X_j\text{ s.t. }j\notin N_i\}$.
\begin{theorem}[Adapted from Theorem 3.6 in \cite{ross2011fundamentals}]
\label{thm:stein}
Let $d_K$ be the Kolmogorov metric, so that for random variables $X$ and $Y$ 
$$d_K(X,Y)=\sup_{x\in\mathbb{R}}|F_X(x)-F_Y(x)|,$$
where $F_X$ (resp., $F_Y$) is the distribution function of $X$ (resp., $Y$).
Let $X_1,\ldots,X_n$ be random variables such that for all $i\in[n]$, 
$\mathbb{E}(X_i^4)<\infty$, 
$\mathbb{E}(X_i)=0$,
$\sigma^2=\text{Var}(\sum_i X_i)$,
and define $W=\sum_i X_i/\sigma$.
Let the collection $(X_1,\ldots,X_n)$ have dependency neighborhoods $N_i$, $i=1,\ldots,n$,
and also define $D:=\max_{i\in[n]}|N_i|$.
Then for $Z$ a standard normal random variable
$$d_K(W,Z)\leq \sqrt{(2/\pi)^{1/2}\left(\frac{D^2}{\sigma^3}\sum_{i=1}^n\mathbb{E}|X_i|^3+\frac{\sqrt{28}D^{3/2}}{\sqrt{\pi}\sigma^2}\sqrt{\sum_{i=1}^n \mathbb{E}(X_i^4)}\right)}$$
\end{theorem}
Recalling that $\sigma$ is the permutation associated with $P(P^*)^T$, define
    $$Y_{h\ell}:=\big(\sum_{i,j}S_i^{(j)}[h,\ell]\big)(A[\sigma(h),\sigma(\ell) ]-A[h,\ell]).$$
    Note that the maximum size of the dependency neighborhoods for the $(Y_{h\ell})$'s is at most 2 (i.e., $D$ in Theorem \ref{thm:stein} is 2).
Let 
$$\displaystyle \alpha_{hl} = \sum_{i,j}S_i^{(j)}[h,l]$$ and 
$$\beta_{hl} = \left(A[\sigma(h),\sigma(l)]-A[h,l]\right),$$ so that $Y_{hl} = \alpha_{hl}\beta_{hl}$. 
It is immediate that conditioning on $B^{(i)},i=1,2$, we have $\{\alpha_{hl}\}_{h,l}$ is  independent of $\{\beta_{hl}\}_{h,l}$.
Below, we will implicitly condition on $B^{(i)},i=1,2$ in all expectations. 

We define
$$X_{hl} := Y_{hl}-\mathbb{E}(Y_{hl}) = \alpha_{hl}\beta_{hl}-\mathbb{E}(\alpha_{hl})\mathbb{E}(\beta_{hl})$$
We first note:
$$
\begin{aligned}
    \mathbb{E}[X_{hl}^4] &= \mathbb{E}\left(\left[\alpha_{hl}\beta_{hl}-\mathbb{E}(\alpha_{hl})\mathbb{E}(\beta_{hl})\right]^4\right)\\
    &= \mathbb{E}\left(\left(\left[\alpha_{hl}-\mathbb{E}(\alpha_{hl})\right]\left[\beta_{hl} + \mathbb{E}(\beta_{hl})\right] + [\beta_{hl} \mathbb{E}(\alpha_{hl}) - \alpha_{hl} \mathbb{E}(\beta_{hl})]\right)^4\right)\\
    &\leq 2^3\Big( \mathbb{E}([\alpha_{hl}-\mathbb{E}(\alpha_{hl})]^4)\mathbb{E}([\beta_{hl}+\mathbb{E}(\beta_{hl})]^4) + \mathbb{E}([\beta_{hl} \mathbb{E}(\alpha_{hl}) - \alpha_{hl} \mathbb{E}(\beta_{hl})]^4)\Big)\\
    &= 2^3\Big( \mathbb{E}([\alpha_{hl}-\mathbb{E}(\alpha_{hl})]^4)\mathbb{E}([\beta_{hl}+\mathbb{E}(\beta_{hl})]^4) + \mathbb{E}([\beta_{hl} (\mathbb{E}(\alpha_{hl}) - \alpha_{hl}) + \alpha_{hl}(\beta_{hl}- \mathbb{E}(\beta_{hl}))]^4)\Big)\\
    &\leq 2^3\Big( \mathbb{E}([\alpha_{hl}-\mathbb{E}(\alpha_{hl})]^4)\mathbb{E}([\beta_{hl}+\mathbb{E}(\beta_{hl})]^4) + 2^3\Big[ \mathbb{E}(\beta_{hl}^4) \mathbb{E}([ \alpha_{hl}-\mathbb{E}(\alpha_{hl}) ]^4) + \mathbb{E}(\alpha_{hl}^4)\mathbb{E}([\beta_{hl}- \mathbb{E}(\beta_{hl})]^4)\Big]\Big)\\
    &= A_1 \mathbb{E}([ \alpha_{hl}-\mathbb{E}(\alpha_{hl}) ]^4) + B_2 \mathbb{E}(\alpha_{hl}^4)
\end{aligned}$$
where $A_1 = 8\mathbb{E}([\beta_{hl}+\mathbb{E}(\beta_{hl})]^4) + 64 \mathbb{E}(\beta_{hl}^4)$ and $B_2 = 64 \mathbb{E}([\beta_{hl}+E(\beta_{hl})]^4)$.

Note that $\alpha_{hl}$ follows the Poisson-binomial distribution with $m$ independent summands. The $4$-th central moment of the Poisson-binomial distribution can be calculated via its Excess Kurtosis which has magnitude  $O(1/m)$ and its variance which has magnitude of $\sigma^2=O(m)$. 
The 4th central moment therefore has magnitude of $O(1/m)O(m^2) = O(m)$.
The 4th non-central moment of the Poisson binomial distribution is of order $O(m^4)$.
Turning our attention to $\beta$, there are three cases to consider:
\begin{enumerate}
    \item If $B^{(1)}[h,l] = B^{(1)}[\sigma(h), \sigma(l)]$, then we know 
    $$\mathbb{P}(\beta_{hl}=a)=\begin{cases}p^2+(1-p)^2,& a=0\\p(1-p),&a=1\\p(1-p),&a=-1\end{cases}$$ and 
    $\mathbb{E}(\beta_{hl}) = 0; \mathbb{V}(\beta_{hl}) = 2p(1-p)$
    \item If $B^{(1)}[h,l] =1, B^{(1)}[\sigma(h), \sigma(l)]=0$, then we know 
    $$P(\beta_{hl}=a)=\begin{cases}2p(1-p),&
    a=0\\p^2,&a=1\\(1-p)^2,&a=-1\end{cases}$$
       $\mathbb{E}(\beta_{hl}) = 2p-1; \mathbb{V}(\beta_{hl}) = 2p(1-p)$
        \item If $B^{(1)}[h,l] =0, B^{(1)}[\sigma(h), \sigma(l)]=1$, then we know 
    $$P(\beta_{hl}=a)=\begin{cases}2p(1-p),&
    a=0\\p^2,&a=-1\\(1-p)^2,&a=1\end{cases}$$
       $\mathbb{E}(\beta_{hl}) = 1-2p; \mathbb{V}(\beta_{hl}) = 2p(1-p)$
\end{enumerate}
Now, it is clear that $A_1 = 8E[\beta_{hl}+E(\beta_{hl})]^4 + 64 E(\beta_{hl}^4)$ and $B_2 = 64 E[\beta_{hl}+E(\beta_{hl})]^4$ are two constants that do not grow with $m$.  This yields that $E[X_{hl}^4] = O(m^4)$.
Similarly, we can show $E[|X_{hl}|^3] = O(m^3)$.


Next, we have (where $\mathbb{V}$ is shorthand for variance, and we are implicitly conditioning on the $B^{(i)}$'s below)
$$\begin{aligned}
    \mathbb{V}\left(\sum_{\substack{h,l,\,\text{ s.t.} \\\{\sigma(h),\sigma(l)\}\neq\{h,l\}}} X_{hl}\right)&=\mathbb{V}\left(\sum_{\substack{h,l,\,\text{ s.t.} \\\{\sigma(h),\sigma(l)\}\neq\{h,l\}}} Y_{hl}\right)\\
    &=\underbrace{\sum_{\substack{h,l,\,\text{ s.t.} \\\{\sigma(h),\sigma(l)\}\neq\{h,l\}}} \mathbb{V}(Y_{hl})}_{:=V1} \\
    &\hspace{5mm}+ \underbrace{\sum_{\substack{h,l,\,\text{ s.t.} \\\{\sigma(h),\sigma(l)\}\neq\{h,l\}}}\sum_{\substack{h_2,l_2\text{ s.t. }\{h_2,l_2\}\neq \{h,l\}\text{ and }\\ \{\sigma(h_2),\sigma(l_2)\}\neq\{h_2,l_2\}}
    }\text{Cov}(Y_{hl}, Y_{h_2l_2})}_{:=C2}
\end{aligned}$$
Now, as $PP^*$ shuffles exactly $k$ labels, the size of the set 
$\{h,l,\,\text{ s.t. } \{\sigma(h),\sigma(l)\}\neq\{h,l\}\}$
is $\Theta(nk)$.  We then have
\begin{align*}
    V1 &= \sum_{\substack{h,l,\,\text{ s.t.} \\\{\sigma(h),\sigma(l)\}\neq\{h,l\}}}\mathbb{V}(\alpha_{hl}\beta_{hl})\\
    &= \sum_{\substack{h,l,\,\text{ s.t.} \\\{\sigma(h),\sigma(l)\}\neq\{h,l\}}} \mathbb{V}(\alpha_{hl})\mathbb{V}(\beta_{hl}) + [\mathbb{E}(\beta_{hl})]^2\mathbb{V}(\alpha_{hl}) + [\mathbb{E}(\alpha_{hl})]^2\mathbb{V}(\beta_{hl})\\
    &= \sum_{\substack{h,l,\,\text{ s.t.} \\\{\sigma(h),\sigma(l)\}\neq\{h,l\}}} \mathbb{E}(\beta_{hl}^2)\underbrace{\mathbb{V}(\alpha_{hl})}_{=\Theta(m)} + \mathbb{V}(\beta_{hl}) [\mathbb{E}(\alpha_{hl})]^2\\
    &= \Theta(nkm) + 2p(1-p)\sum_{\substack{h,l,\,\text{ s.t.} \\\{\sigma(h),\sigma(l)\}\neq\{h,l\}}}[\mathbb{E}(\alpha_{hl})]^2
\end{align*}
Next, we have
\begin{align*}
    C2 &= \sum_{\substack{h,l,\,\text{ s.t.} \\\{\sigma(h),\sigma(l)\}\neq\{h,l\}}}\left\{\text{Cov}\left[\alpha_{hl}\beta_{hl}, \alpha_{\sigma^{-1}(h)\sigma^{-1}(l)}\beta_{\sigma^{-1}(h)\sigma^{-1}(l)}\right] + \text{Cov}\left[\alpha_{hl}\beta_{hl}, \alpha_{\sigma(h)\sigma(l)}\beta_{\sigma(h)\sigma(l)}\right]\right\}\\
    &=-\sum_{\substack{h,l,\,\text{ s.t.} \\\{\sigma(h),\sigma(l)\}\neq\{h,l\}}}\{ \mathbb{E}(\alpha_{hl})\mathbb{E}(\alpha_{\sigma^{-1}(h)\sigma^{-1}(l)}) \mathbb{V}(A[h,l]) + \mathbb{E}(\alpha_{hl})\mathbb{E}(\alpha_{\sigma(h)\sigma(l)}) \cdot \mathbb{V}(A[\sigma(h),\sigma(l)])\}\\
    &= -2p(1-p)\sum_{\substack{h,l,\,\text{ s.t.} \\\{\sigma(h),\sigma(l)\}\neq\{h,l\}}} \mathbb{E}(\alpha_{hl})\mathbb{E}(\alpha_{\sigma(h)\sigma(l)})
\end{align*}
Combining, we then see
\begin{align*}
    \mathbb{V}&\left(\sum_{\substack{h,l,\,\text{ s.t.} \\\{\sigma(h),\sigma(l)\}\neq\{h,l\}}} X_{hl}\right) = \Theta(nkm)+p(1-p)\sum_{\substack{h,l,\,\text{ s.t.} \\\{\sigma(h),\sigma(l)\}\neq\{h,l\}}}\left(\mathbb{E}(\alpha_{hl})-\mathbb{E}(\alpha_{\sigma(h)\sigma(l)})\right)^2\\
\end{align*}
Now, $\alpha_{hl}$ follows the Poisson-Binomial distribution, and
$$\mathbb{E}(\alpha_{hl})=m_1\left((1-2p)B^{(1)}[h,l]+p \right)+m_2\left((1-2p)B^{(2)}[h,l]+p \right),$$
and so
$$\mathbb{E}(\alpha_{\sigma(h)\sigma(l)})-\mathbb{E}(\alpha_{hl})=(1-2p)\left(m_1\left(B^{(1)}[\sigma(h),\sigma(l)]-B^{(1)}[h,l]\right)+m_2\left(B^{(2)}[\sigma(h),\sigma(l)]-B^{(2)}[h,l]\right)\right).$$

Without loss of generality, we will consider $P^*=I_n$ below (this is done to simply ease notation).
By assumption, we have that 
\begin{align}
\label{eq:diff}
    \frac{ h(B^{(2)},B^{(1)},P)}{ -h(B^{(1)},B^{(1)},P)}> \frac{m_1(1-2p)}{m_2(1-2p)}=\frac{m_1}{m_2},
\end{align}
where we recall
$$
h(B^{(i)},B^{(j)},P)=\text{tr}(B^{(i)} P B^{(j)} P^T)-\text{tr}(B^{(i)}B^{(j)})
.$$
Note that the possible values of $\left(B^{(1)}[\sigma(h),\sigma(l)]-B^{(1)}[h,l]\right)$ and $\left(B^{(2)}[\sigma(h),\sigma(l)]-B^{(2)}[h,l]\right)$ are $-1, 0$ or 1.
A key term in the variance computation above is
\begin{align}
&p(1-p)\sum_{\substack{h,l,\,\text{ s.t.} \\\{\sigma(h),\sigma(l)\}\neq\{h,l\}}}\left(\mathbb{E}(\alpha_{hl})-\mathbb{E}(\alpha_{\sigma(h)\sigma(l)})\right)^2\notag\\
&=
p(1-p)(1-2p)^2\sum_{\substack{h,l,\,\text{ s.t.} \\\{\sigma(h),\sigma(l)\}\neq\{h,l\}}}\left(m_1\left(B^{(1)}[\sigma(h),\sigma(l)]-B^{(1)}[h,l]\right)+m_2\left(B^{(2)}[\sigma(h),\sigma(l)]-B^{(2)}[h,l]\right)\right)^2\label{eq:Varterm}
\end{align}
We desire (for the application of Stein's method in Theorem \ref{thm:stein}) that this term is $\omega(m^2(nk)^{2/3})$.  When will this be the case?

For each $x\in\{0,1\}^4$, let 
$$N_x:=\left|\left\{\,\{h,\ell\}\in\binom{V}{2}\text{ s.t. }\bigg(B^{(1)}[\sigma(h),\sigma(l)],B^{(1)}[h,l],B^{(2)}[\sigma(h),\sigma(l)],B^{(2)}[h,l]\bigg)=x\right\}\right| $$
Note that, by parity, we have 
\begin{align*}
    N_{0110}+N_{0111}+N_{0100}+N_{0101}&=N_{1010}+N_{1011}+N_{1000}+N_{1001}\\
    N_{0001}+N_{1101}+N_{1001}+N_{0101}&=N_{0010}+N_{1110}+N_{1010}+N_{0110}
\end{align*}
Equation \ref{eq:diff} is then equivalent to 
$$ m_2(N_{0110}+N_{1110}-N_{0101}-N_{1101})>m_1(N_{0110}+N_{0111}+N_{0100}+N_{0101}).$$
This then implies
\begin{align}
    &\frac{m_2}{2}(N_{0110}+N_{1110}-N_{0101}-N_{1101})+\frac{m_2}{2}(N_{1001}+N_{0001}-N_{1010}-N_{0010})\notag\\
    &\hspace{10mm}>\frac{m_1}{2}(N_{0110}+N_{0111}+N_{0100}+N_{0101})+\frac{m_1}{2}(N_{1010}+N_{1011}+N_{1000}+N_{1001})\notag\\
    &\Leftrightarrow \frac{m_2}{2}(N_{0110}+N_{1110}+N_{1001}+N_{0001})>\frac{m_1}{2}(N_{0110}+N_{0111}+N_{0100}+N_{0101})\notag\\
    &\hspace{70mm}+\frac{m_1}{2}(N_{1010}+N_{1011}+N_{1000}+N_{1001})\notag\\
    &\hspace{70mm}+\frac{m_2}{2}(N_{0101}+N_{1101}+N_{1010}+N_{0010})\label{eq:diff2}
\end{align}
Note that in Eq. \ref{eq:Varterm}, each 
\begin{align*}
    N_{1010}&\text{ term contributes }(m_1+m_2)^2;\quad
    N_{0101}\text{ term contributes }(m_1+m_2)^2;\\
    N_{1001}&\text{ term contributes }(m_1-m_2)^2;\quad
    N_{0110}\text{ term contributes }(m_1-m_2)^2;\\
    N_{0010}&\text{ term contributes }m_2^2;\quad
    N_{1110}\text{ term contributes }m_2^2;\\
    N_{0001}&\text{ term contributes }m_2^2;\quad
    N_{1101}\text{ term contributes }m_2^2;\\
    N_{1011}&\text{ term contributes }m_1^2;\quad
    N_{1000}\text{ term contributes }m_1^2;\\
    N_{0111}&\text{ term contributes }m_1^2;\quad
    N_{0100}\text{ term contributes }m_1^2.
\end{align*}
We consider the following cases:
\begin{itemize}
    \item[i.]{$\mathbf{|m_1-m_2|=o(m):}$} In this case the $N_{1001}$ and $N_{0110}$ terms contribute minimally  (i.e., of order $o(m^2)$ and not of order $m^2$) to Eq. \ref{eq:Varterm}.
    In order for Eq. \ref{eq:Varterm} to be of order $\omega(m^2(nk)^{2/3})$ it is necessary and sufficient for at least one of 
$$N_{1010},\,
N_{0101},\,
N_{0010},\,
N_{1110},\,
N_{0001},\,
N_{1101},\,
N_{1011},\,
N_{1000},\,
N_{0111},\,
N_{0100}$$
to be $\omega((nk)^{2/3})$,
which, by Eq. \ref{eq:diff2}, is equivalent to 
  $$N_{1110}+N_{0001}=\omega((nk)^{2/3}).$$
%
\item[ii.]{$\mathbf{m_1,m_2=\Theta(m),\,|m_1-m_2|=\Theta(m):}$} In this case, all terms contribute meaningfully (i.e., order $m^2$) to Eq. \ref{eq:Varterm}.
        If $m=\omega(1)$, then in order for Eq. \ref{eq:Varterm} to be of order $\omega(m^2(nk)^{2/3})$ it is necessary and sufficient for at least one of 
    $$N_{1010},\,
    N_{0101},\,
      N_{1001},\,
    N_{0110},\,
    N_{0010},\,
    N_{1110},\,
    N_{0001},\,
    N_{1101},\,
         N_{1011},\,
              N_{1000},\,
                   N_{0111},\,
                        N_{0100},
    $$
    to be $\omega((nk)^{2/3})$,
    which, by Eq. \ref{eq:diff2}, is equivalent to 
      $$N_{1110}+N_{0001}+N_{1001}+N_{0110}=\omega((nk)^{2/3}).$$
    \item[iii.]{$\mathbf{m_2/m_1=\omega(1):}$} In this case the 
    $N_{1011}$,
    $N_{1000}$,
    $N_{0111}$,
    and $N_{0100}$ terms contribute minimally  (i.e., of order $m_1^2\ll m^2$) to Eq. \ref{eq:Varterm}.
        If $m=\omega(1)$, then in order for Eq. \ref{eq:Varterm} to be of order $\omega(m^2(nk)^{2/3})$ it is necessary and sufficient for at least one of 
    $$N_{1010},\,
    N_{0101},\,
    N_{1001},\,
    N_{0110},\,
    N_{0010},\,
    N_{1110},\,
    N_{0001},\,
    N_{1101},
    $$
    to be $\omega((nk)^{2/3})$,
    which, by Eq. \ref{eq:diff2}, is equivalent to 
      $$N_{1110}+N_{0001}+N_{1001}+N_{0110}=\omega((nk)^{2/3}).$$
\end{itemize}
If the conditions above hold, we have 
\begin{align*}
    \mathbb{V}&\left(\sum_{\substack{h,l,\,\text{ s.t.} \\\{\sigma(h),\sigma(l)\}\neq\{h,l\}}} X_{hl}\right) = \Theta(nkm)+p(1-p)\sum_{\substack{h,l,\,\text{ s.t.} \\\{\sigma(h),\sigma(l)\}\neq\{h,l\}}}\left(\mathbb{E}(\alpha_{hl})-\mathbb{E}(\alpha_{\sigma(h)\sigma(l)})\right)^2=\Theta(nkm) + \omega(m^2(nk)^{2/3})
\end{align*}
In this case, the bound in Stein's method becomes( where  
$\sum_*$ is shorthand for $\sum_{\substack{h,l,\,\text{ s.t.} \\\{\sigma(h),\sigma(l)\}\neq\{h,l\}}}$)
and
$$W=\sum_* X_{hl}/\sqrt{\mathbb{V}_B(\sum_* X_{hl})},
$$
\begin{align*}
d_K(W,Z)&\leq \sqrt{\frac{O(nkm^3)}{\Theta((nkm)^{3/2}) + \omega(m^3nk)}+\frac{O((nk)^{1/2}m^2)}{\Theta(nkm) + \omega(m^2(nk)^{2/3})}}=o(1)
\end{align*}
as desired.
In the event that none of the growth conditions outlined above for the $N_{ijkl}$'s hold, then 
$$\mathbb{V}_B(\sum_* X_{hl})=\Omega(nkm)$$ and we can bound $d_K(W,Z)$ via
\begin{align*}
d_K(W,Z)&\leq \sqrt{\frac{O(nkm^3)}{\Omega((nkm)^{3/2}) }+\frac{O((nk)^{1/2}m^2)}{\Omega(nkm)}}=\sqrt{O\left(\frac{m^{3/2}}{(nk)^{1/2}}\right)+O\left(\frac{m}{(nk)^{1/2}}\right)}
\end{align*}
and this bound is $o(1)$ when $nk\gg m^3$ as desired.
\end{proof}

\subsection{Proof of Corollary \ref{cor:cor1}}
\label{app:cor1}

By the normal convergence in Theorem \ref{thm:thm1}, we have that (where $Z\sim$N(0,1))
\begin{align*}
\mathbb{P}(f(P)-f(P^*)> 0)&=\mathbb{P}\left(
\frac{f(P)-f(P^*)-\mathbb{E}_B(f(P)-f(P^*))}{\sqrt{\text{Var}_B(f(P)-f(P^*))}} > \frac{-\mathbb{E}_B(f(P)-f(P^*))}{\sqrt{\text{Var}_B(f(P)-f(P^*))}}\right)\\
&\geq  \mathbb{P}\left(
\frac{f(P)-f(P^*)-\mathbb{E}_B(f(P)-f(P^*))}{\sqrt{\text{Var}_B(f(P)-f(P^*))}} > 0\right)\\
&\rightarrow \mathbb{P}(Z>0)=1/2.
\end{align*}
For part ii., let $\epsilon>0$ fixed.  
We have that for any constant $C>0$,
\begin{align*}
\mathbb{P}(f(P)-f(P^*)> 0)&=\mathbb{P}\left(
\frac{f(P)-f(P^*)-\mathbb{E}_B(f(P)-f(P^*))}{\sqrt{\text{Var}_B(f(P)-f(P^*))}} > \frac{-\mathbb{E}_B(f(P)-f(P^*))}{\sqrt{\text{Var}_B(f(P)-f(P^*))}}\right)\\
&\geq  \mathbb{P}\left(
\frac{f(P)-f(P^*)-\mathbb{E}_B(f(P)-f(P^*))}{\sqrt{\text{Var}_B(f(P)-f(P^*))}} > \frac{-Cm\sqrt{n\xi\log n}}{\sqrt{n\xi m^2}}\right)\\
&\geq  \mathbb{P}\left(
\frac{f(P)-f(P^*)-\mathbb{E}_B(f(P)-f(P^*))}{\sqrt{\text{Var}_B(f(P)-f(P^*))}} > -C\sqrt{\log n}\right)
\end{align*}
For $n$ sufficiently large, this last term is bounded below by (where $\Phi(x)=\mathbb{P}(Z\leq x)$)
\begin{align*}
\mathbb{P}&\left(
\frac{f(P)-f(P^*)-\mathbb{E}_B(f(P)-f(P^*))}{\sqrt{\text{Var}_B(f(P)-f(P^*))}} > -\Phi^{-1}(1-\epsilon)\right)\\
&\rightarrow \Phi(\Phi^{-1}(1-\epsilon))=1-\epsilon.
\end{align*}
As $\epsilon$ was arbitrary, letting it go to $0$ finishes the proof.

\subsection{Proof of Lemma \ref{prop:Q}}
\label{app:pf2}
    Suppose such $P$ matrix exists, for any graph $B^{(j)}$, where $j=2,3,\ldots,m$, we 
    consider the matching objective function $$\|P^T\mathbb{E}(B^{(1)})P - \mathbb{E}(B^{(j)})\|_F^2=\|P^\top U R^{(1)} U^TP - U R^{(j)}U^T\|_F^2.$$ 
    We can lift $U$ and $R$'s to $\tilde{U}$ and $\tilde{R}^{(j)}$ such that $\tilde{U}$ is an orthogonal matrix, $\tilde{R}^{(j)}$'s are still diagonal matrices, and $\tilde{U}\tilde{R}^{(j)}\tilde{U}^T = \mathbb{E}(B^{(j)})$ for all $j$. Therefore we know
    \begin{align*}
    -\|P^T U R^{(1)} U^TP - U R^{(j)}U^T\|_F^2 =& -\|P^T \tilde{U} \tilde{R}^{(1)} \tilde{U}^TP - \tilde{U} \tilde{R}^{(j)}\tilde{U}^T\|_F^2\\
	=& -\|P^T\tilde{U} \tilde{R}^{(1)} \tilde{U}P^T\|^2_F - \|\tilde{U} \tilde{R}^{(j)}\tilde{U}^T\|_F^2+2\operatorname{tr}(P^T \tilde{U} \tilde{R}^{(1)} \tilde{U}^TP\tilde{U} \tilde{R}^{(j)}\tilde{U}^T)\\
	=& 2\operatorname{tr}(\tilde{R}^{(1)}X\tilde{R}^{(j)}X^T) - K
    \end{align*}
    where $X =  \tilde{U}^TP\tilde{U}$, and $K = -\|\tilde{R}^{(1)}\|_F^2 - \|\tilde{R}^{(j)}\|_F^2 \in \mathbb{R}$ is independent of $P$.
    
    For general $X\in\mathbb{R}^{d\times d}$, define the matrix functional $f_2(X) = \operatorname{tr}(\tilde{R}^{(1)}X\tilde{R}^{(j)}X^T)$.
    Letting $\tilde Q=Q\oplus \mathbf{0}_{n-d}$ (where $\mathbf{0}_{n-d}$ is the $(n-d)\times(n-d)$ matrix of all $0$'s), we we have that $f_2(\tilde Q) > f_2(I)=f_2(\tilde{U} I\tilde{U}^T)$ by assumption.
    Further define the functional 
    $$g_2(X) = \sqrt{\operatorname{tr}\left({\left(\tilde{R}^{(1)}\right)}^2X{\left(\tilde{R}^{(j)}\right)}^2X^T\right)}.$$ 
 The diagonal elements of each $R^{(j)}$ are nonnegative, and by the $\ell1-\ell2$ norm inequality, 
    we have that $f_2(\tilde Q)\geq g_2(\tilde Q)$.
    Let $W = U^T P U\oplus \mathbf{0}_{n-d}$, and define 
    $$\epsilon=\|U^T P U-Q\|_F=\|W-\tilde Q\|_F.$$
Recall our assumption that 
      \begin{align*}
        Q&\in\text{argmin}_{V\in\Pi_d}\|R^{(1)}-VR^{(j)}V^T\|_F\\
        I_d&\notin\text{argmin}_{V\in\Pi_d}\|R^{(1)}-VR^{(j)}V^T\|_F,
    \end{align*} 
    Now, we know that 
    \begin{align*}
        f_2(W)&=\operatorname{tr}(\tilde{R}^{(1)}W\tilde{R}^{(j)}W^T)\\
        &=\operatorname{tr}(\underbrace{\tilde U\tilde{R}^{(1)}\tilde U^T}_{\mathbb{E}(B^{(1)})}\underbrace{ P\tilde U\tilde{R}^{(j)}\tilde U^T P^T}_{P\mathbb{E}(B^{(j)})P^T})
    \end{align*}
    As both $\mathbb{E}(B^{(1)})$ and $P\mathbb{E}(B^{(j)})P^T$ are Hermition with respective eigenvalues the diagonal entries of $R^{(1)}$ and $R^{(j)}$, we have that 
        \begin{align*}
        \operatorname{tr}(\mathbb{E}(B^{(1)})P\mathbb{E}(B^{(j)})P^T)\leq f_2(Q)
    \end{align*}
    as $Q$ sorts the eigenvalues of $\mathbb{E}(B^{(1)})$ and $P\mathbb{E}(B^{(j)})P^T$ to both be in non-decreasing order; see Theorem 1 in \cite{hoffman2003variation}.
    Similarly $g_2(W)\leq g_2(\tilde Q)$. Now we consider the mean value theorem (MVT) applied to the function $f_2$:
    By the multivariate MVT, we know there is a point $c\tilde Q+(1-c)W$ where $c\in (0,1)$ such that 
    $$f_2(\tilde Q)-f_2(W) = \left(\operatorname{vec}(\nabla f_2(c\tilde Q+(1-c)W))\right)^T \operatorname{vec}(\tilde Q-W)$$
    Plugging in $\nabla f_2(X) = 2\tilde{R}^{(1)}X\tilde{R}^{(j)}$, we get
    \begin{align*}
        f_2(\tilde Q)-f_2(W) &= 2\left(\operatorname{vec}(\tilde{R}^{(1)}(c\tilde Q+(1-c)W)\tilde{R}^{(j)})\right)^T \operatorname{vec}(\tilde Q-W)\\
        &= 2\left[(\tilde{R}^{(1)}\otimes\tilde{R}^{(j)})\operatorname{vec}(c\tilde Q+(1-c)W)\right]^T\operatorname{vec}(\tilde Q-W)\\
        &\leq 2\left\|\operatorname{vec}(c\tilde Q+(1-c)W)^T(\tilde{R}^{(1)}\otimes\tilde{R}^{(j)})\right\|_2\|\operatorname{vec}(\tilde Q-W)\|_2\\
        &=2\|\tilde{R}^{(1)}(c\tilde Q+(1-c)W )\tilde{R}^{(j)}\|_F \|\tilde Q-W\|_F\\
        &\leq 2\varepsilon \left(c\|\tilde{R}^{(1)}\tilde Q\tilde{R}^{(j)}\|_F + (1-c)\|\tilde{R}^{(1)}W\tilde{R}^{(j)}\|_F\right)\\
        &\leq 2\varepsilon \|\tilde{R}^{(1)}\tilde Q\tilde{R}^{(j)}\|_F\\
        &=2\varepsilon \sqrt{\operatorname{tr}\left({\left(\tilde{R}^{(1)}\right)}^2\tilde Q{\left(\tilde{R}^{(j)}\right)}^2\tilde Q^T\right)}\\
        &= 2\varepsilon g_2(\tilde Q)\\
        &\leq 2\varepsilon f_2(\tilde Q)
    \end{align*}
    Thus we conclude $f_2(W)\geq (1-2\varepsilon) f_2(\tilde Q)$, which implies (by the assumption on $P$)
  \begin{align*}
      \operatorname{tr}(P^T \mathbb{E}&(B^{(1)})P\mathbb{E}(B^{(j)}))=\operatorname{tr}(P^T \tilde{U} \tilde{R}^{(1)} \tilde{U}^TP\tilde{U} \tilde{R}^{(j)}\tilde{U}^T)=\operatorname{tr}(W^T \tilde{R}^{(1)} W \tilde{R}^{(j)})\geq (1-2\varepsilon) \operatorname{tr}(\tilde Q^T \tilde{R}^{(1)} \tilde Q \tilde{R}^{(j)})
   \\
    &>\operatorname{tr}(\tilde{R}^{(1)}\tilde{R}^{(j)})=\operatorname{tr}( \tilde{U} \tilde{R}^{(1)} \tilde{U}^T\tilde{U} \tilde{R}^{(j)}\tilde{U}^T)=\operatorname{tr}( \mathbb{E}(B^{(1)})\mathbb{E}(B^{(j)}))
  \end{align*}
  as desired.

\subsection{Additional computational details}
\label{app:addcomp}
\subsubsection{Proof of Proposition \ref{prop:33} McDiarmid Concentration}
Fix  $\xi$.
From Eq. \ref{eq:fdiff} we can see that $f(P)-f(P^\ast)$ is a function of $\Theta(n\xi)$ independent random variables
$$\left\{A[h,\ell], X[h,\ell]:=\left(\sum_{i,j}S^{(j)}_i[h,\ell]\right)\right\}_{\{\sigma(h),\sigma(\ell)\}\neq\{h,\ell\}}.$$
A single change in one of these variables can change the value of $f(P)-f(P^\ast)$ by at most $O(m)$.
Suppose that $\mathbb{E}_B(f(P)-f(P^*))<0$ holds, then by McDiarmid Inequality, we know for any fixed $\xi < n$ and $P$ such that $P(P^*)^T\in \Pi_{n,\xi}$, we have
\begin{align*}
    \mathbb{P}\left(\left[f(P)-f(P^\ast)\right]\geq 0\right)\leq& \mathbb{P}\left(\big|\left[f(P)-f(P^\ast)\right]-\mathbb{E}_B(f(P)-f(P^*))\big|\geq \mathbb{E}_B(f(P)-f(P^*))\right)\\
    \leq & 2\exp{\left(-\frac{\left[\mathbb{E}_B(f(P)-f(P^*))\right]^2}{O(n\xi m^2)}\right)}\\
    = & 2\exp\left(-\omega(\xi  \log n)\right)  
\end{align*}
A union over such $P$ and $\xi$, we get:
\begin{align*}
\mathbb{P}(\{P^*\}\notin \text{argmin}_P\|C-PRP^T\|_F)&=\mathbb{P}\left\{\exists \xi<n, P\in\Pi_n\text{ s.t. }P(P^*)^T\in \Pi_{n,\xi}:\left[f(P)-f(P^*)\right]\geq 0\right\}\\
&\leq\sum_\xi e^{O(\xi\log(n))} \cdot 2e^{-\omega(\xi \log n)}\\
&=2e^{-\omega( \log n)}=o(1)
\end{align*}
as desired.


\subsubsection{Derivation of Eq. \ref{eq:cmg2}}
\label{app:eq7}
For any graph $B\in\mathcal{G}_n$, let $\bar B$ denote the complement network in $\mathcal{G}_n$.
By linearity of the expectation and the trace, combined with Eq. \ref{eq:needfor5} and the assumptions that all $p_j$ identically equal $p$, we have (where $P^{\bigstar} := P(P^\ast)^T$, $J$ is the hollow $n\times n$ matrix with all off-diagonal entries identically equal to 1 so that $\bar B=J-B$)
\begin{align*}
    \mathbb{E}_B(\text{tr}(C^{(j)}PRP^T))=&\mathbb{E}_B(\text{tr}(S_i^{(j)}PRP^T))\\
    =& (1-p)^2\text{tr}(B^{(j)} P(P^*)^T B^{(1)} P^*P^T)+ p(1-p)\,\text{tr}(\bar B^{(j)} P(P^*)^T B^{(1)} P^*P^T) \\
    &+ (1-p)p\,\text{tr}(B^{(j)} P(P^*)^T \bar B^{(1)} P^*P^T)+ p^2\text{tr}(\bar B^{(j)} P(P^*)^T \bar B^{(1)} P^*P^T)\\
    =&\tr\left(B^{(j)}P^\bigstar B^{(1)}(P^\bigstar)^T\right) -2p\, \tr\left(B^{(j)}P^\bigstar B^{(1)}(P^\bigstar)^T\right) \\
    &+ p^2\tr\left(B^{(j)}P^\bigstar B^{(1)}(P^\bigstar)^T\right) 
    + p\,\tr\left(JP^\bigstar B^{(1)}(P^\bigstar)^T\right)\\
    & -p\,\tr\left(B^{(j)}P^\bigstar B^{(1)}(P^\bigstar)^T\right) 
    - p^2 \tr\left(JP^\bigstar B^{(1)}(P^\bigstar)^T\right)\\
    &+ p^2\tr\left(B^{(j)}P^\bigstar B^{(1)}(P^\bigstar)^T\right) + p\,\tr\left(B^{(j)}P^\bigstar J(P^\bigstar)^T\right)\\
    & -p\,\tr\left(B^{(j)}P^\bigstar B^{(1)}(P^\bigstar)^T\right)- p^2 \tr\left(B^{(j)}P^\bigstar J(P^\bigstar)^T\right) \\
    & + p^2\tr\left(B^{(j)}P^\bigstar B^{(1)}(P^\bigstar)^T\right)+p^2\tr(JP^\bigstar J(P^\bigstar)^T)\\
    &-p^2\tr\left(B^{(j)}P^\bigstar J(P^\bigstar)^T\right) - p^2 \tr\left(JP^\bigstar B^{(1)}(P^\bigstar)^T\right) + p^2\tr\left(B^{(j)}P^\bigstar B^{(1)}(P^\bigstar)^T\right)\\
    =&(1-2p)^2\tr\left(B^{(j)}P^\bigstar B^{(1)}(P^\bigstar)^T\right) + (p-2p^2)\left(\|B^{(1)}\|_F^2+ \|B^{(j)}\|_F^2\right)+p^22\binom{n}{2}
\end{align*}
We note that the identity $\tr\left(JP^\bigstar B^{(j)}(P^\bigstar)^T\right) = \tr(JB^{(j)}) = \tr\left((B^{(j)})^2\right) = \|B^{(j)}\|_F^2$ was used above.
We then get
\begin{align*}
    \mathbb{E}_B&(\text{tr}(C^{(1)}P^*R(P^*)^T))-
    \mathbb{E}_B(\text{tr}(C^{(i)}PRP^T))\\
    &=(1-2p)^2\tr\left(B^{(1)} B^{(1)}\right) + (p-2p^2)\|B^{(1)}\|_F^2 + (p-2p^2)\|B^{(1)}\|_F^2 \\
    &\hspace{5mm}- (1-2p)^2\tr\left(B^{(j)}P^\bigstar B^{(1)}(P^\bigstar)^T\right) - (p-2p^2)\|B^{(1)}\|_F^2 - (p-2p^2)\|B^{(j)}\|_F^2\\
    &=[(1-2p)^2+p-2p^2]\|B^{(1)}\|^2_F -(p-2p^2)\|B^{(j)}\|^2_F -
    (1-2p)^2\text{tr}(B^{(i)}P(P^*)^TB^{(1)}P^*P^T)\\
    &=(1-p)(1-2p)\|B^{(1)}\|_F^2-p(1-2p)\|B^{(i)}\|_F^2 - (1-2p)^2\text{tr}(B^{(i)}P(P^*)^TB^{(1)}P^*P^T)
\end{align*}
as desired.

\subsubsection{Proof of Theorem \ref{thm:secIV}}
\label{app:eq71}
Write 
$$X_{i,P}=\sum_{h\ell} \left(C^{(1)}[h,\ell]-((P^\bigstar)^TC^{(i)}P^\bigstar)[h,\ell] \right)A[\ell,h]$$
is a sum of $O(n^2)$ independent random variables---the $(C^{(1)}[h,\ell]-((P^\bigstar)^TC^{(i)}P^\bigstar)[h,\ell])A[\ell,h]$'s---which are all bounded in $[-2,2]$.
Hoeffding's inequality then yields
\begin{align*}
\mathbb{P}(X_{i,P}\leq 0)&\leq \mathbb{P}(|X_{i,P}-\mathbb{E}X_{i,P}|\geq \mathbb{E}X_{i,P})\\
&\leq 2\text{exp}\left\{-\frac{2(\mathbb{E}X_{i,P})^2}{16n^2}
\right\}\\
&\leq 2\text{exp}\left\{-\omega(\xi\log(n))
\right\}
\end{align*}
Then 
\begin{align*}
\mathbb{P}(\exists i\in[k]\setminus \{1\},P\in\Pi_n\text{ s.t.} X_{i,P}\leq 0)&\leq \sum_{i=2}^k \sum_{P\text{ s.t. }P^\bigstar\in\Pi_{n,\xi}}
2\text{exp}\left\{-\omega(\xi\log(n))\right\}\\
&\leq 2k \sum_{\xi}
\text{exp}\{O(\xi\log(n))\}
\text{exp}\left\{-\omega(\xi\log(n))\right\}\\
&=\text{exp}\{-\omega(\log(n))\}
\end{align*}
as desired.

\subsection{Additional experiments and figures}
\label{app:addexp}
\subsubsection{ER p=0.5}
\begin{figure*}[t!]
    \centering
    \includegraphics[width=\textwidth]{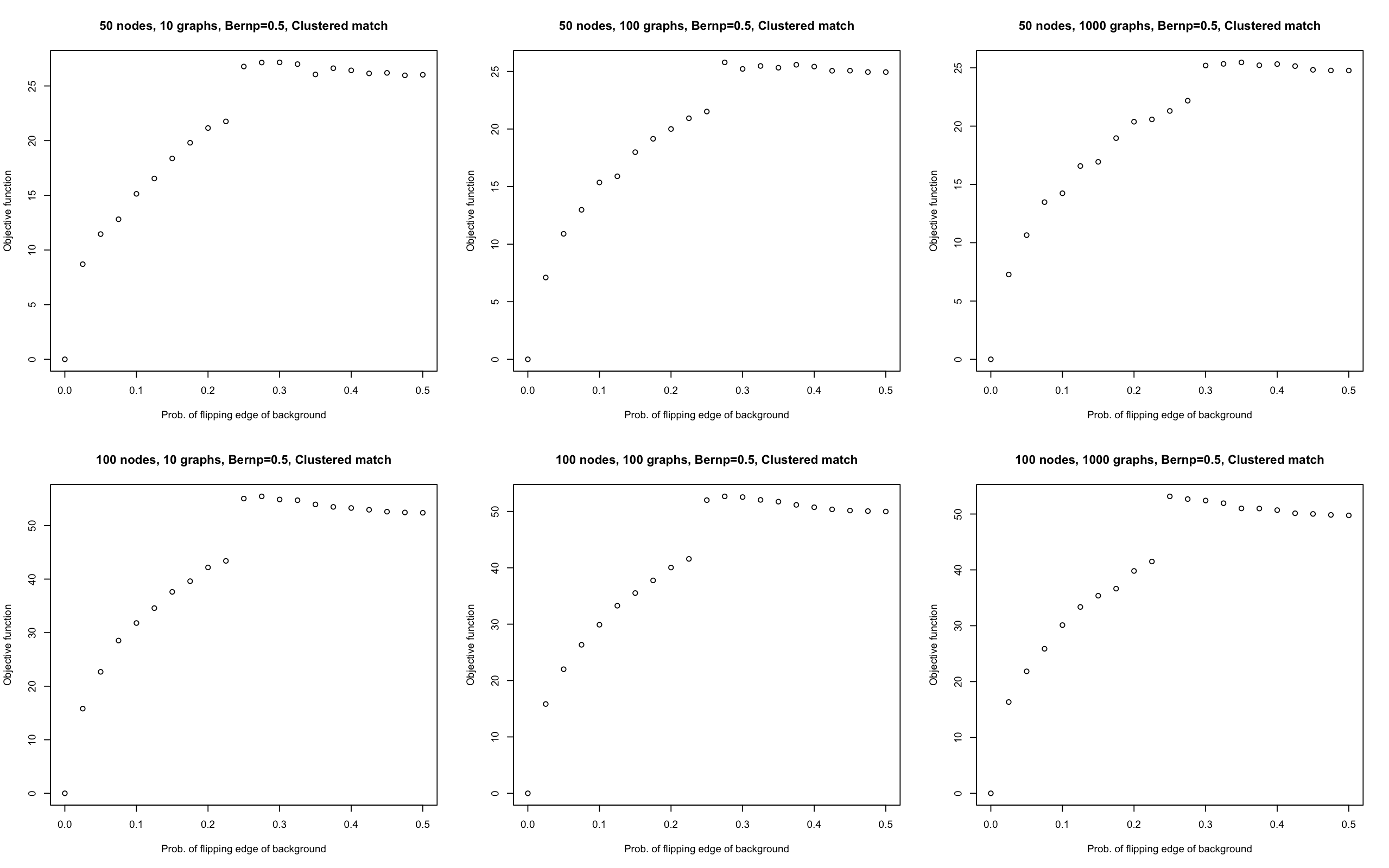}
    \caption{With a single background $B\sim\operatorname{ER}(n,0.5)$, we consider $A,S^{(1)}_i\stackrel{i.i.d.}{\sim}\operatorname{BF}(B,q)$, and we match $A$ (i.e., $P^*=I_n$) to $C$
    using \texttt{SGM} with 5 seeds.
    Varying the number of nodes $n$ ($n=50$ in the top panels, and $n=100$ in the bottom panels), and the number of in-sample graphs ($m=10$ in the left panels, $m=100$ in the middle panels, and $m=1000$ in the right panels), we plot the \texttt{SGM} objective function value $f=\|A-PCP^T\|_F$ versus the value of the edge perturbation parameter $q$, averaged over 10 Monte Carlo iterates.}
    \label{fig:er1}
    \end{figure*}

In this section, we include the results and output of additional experiments.  We first display Table \ref{tab:er05} and Figure \ref{fig:er1} displaying matching accuracy and matching objective function for the ER($n,p=0.5$) single background setting.
\subsubsection{ER p=0.3}
\begin{figure*}[t!]
    \centering
    \includegraphics[width=\textwidth]{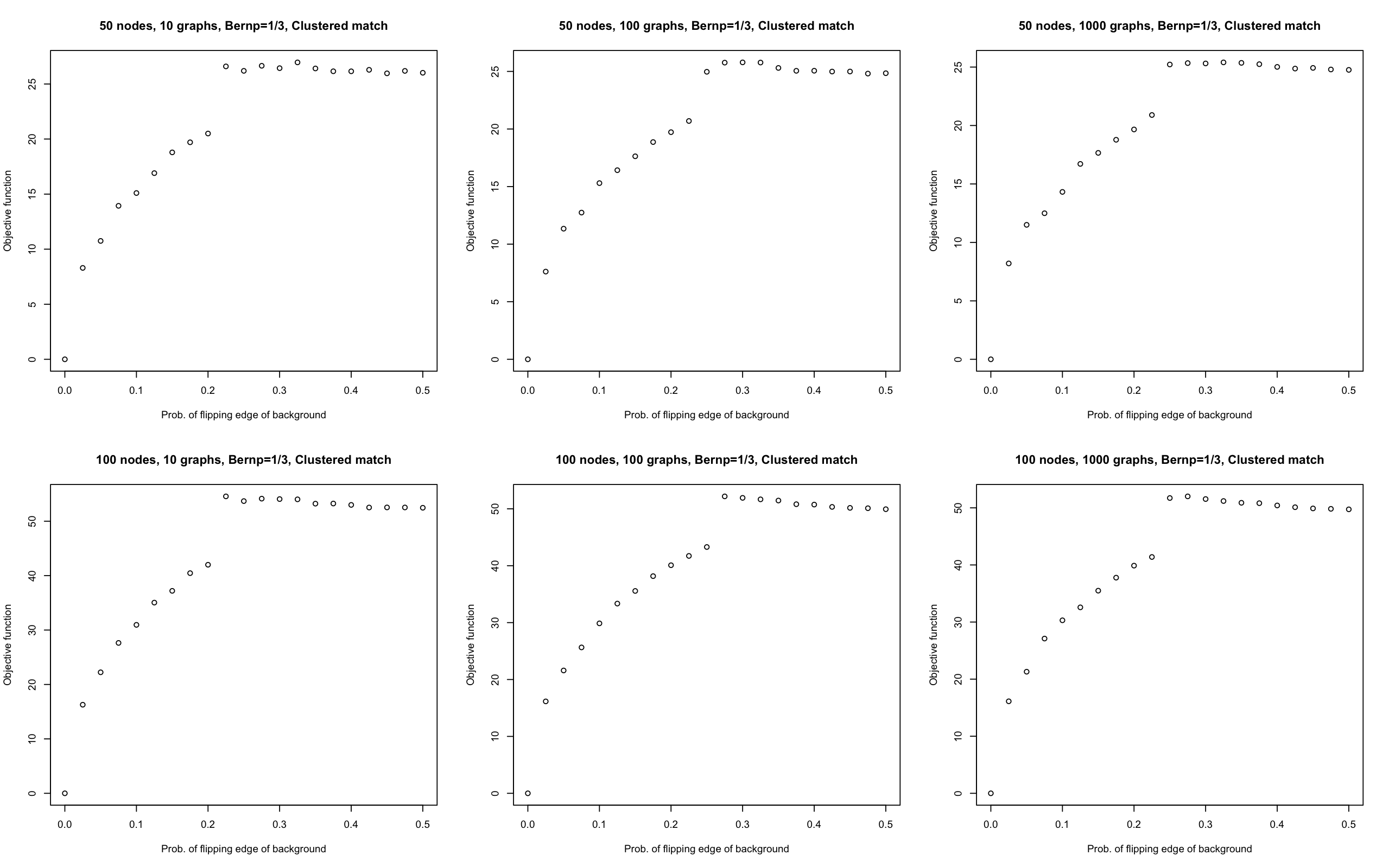}
    \caption{With a single background $B\sim\operatorname{ER}(n,0.3)$, we consider $A,S^{(1)}_i\stackrel{i.i.d.}{\sim}\operatorname{BF}(B,q)$, and we match $A$ (i.e., $P^*=I_n$) to $C$
    using \texttt{SGM} with 5 seeds.
    Varying the number of nodes $n$ ($n=50$ in the top panels, and $n=100$ in the bottom panels), and the number of in-sample graphs ($m=10$ in the left panels, $m=100$ in the middle panels, and $m=1000$ in the right panels), we plot the \texttt{SGM} objective function value $f=\|A-PCP^T\|_F$ versus the value of the edge perturbation parameter $q$, averaged over 10 Monte Carlo iterates.}
    \label{fig:era1}
    \end{figure*}

{We display Table \ref{tab:era13} and Figure \ref{fig:era1} displaying matching accuracy for the ER($n,p=0.3$) single background setting.}
\begin{table}[b!]
\center
\begin{tabular}{|c|c|c|c|c|c|c|}
\hline
$n$         & 50   & 50   & 50   & 100  & 100  & 100  \\ \hline
$m$         & 10   & 100  & 1000 & 10   & 100  & 1000 \\ \hline
$q=0$     & 1    & 1    & 1    & 1    & 1    & 1    \\ \hline
$q=0.025$ & 1    & 1    & 1    & 1    & 1    & 1    \\ \hline
$q=0.050$ & 1    & 1    & 1    & 1    & 1    & 1    \\ \hline
$q=0.075$ & 1    & 1    & 1    & 1    & 1    & 1    \\ \hline
$q=0.100$ & 1    & 1    & 1    & 1    & 1    & 1    \\ \hline
$q=0.125$ & 1    & 1    & 1    & 1    & 1    & 1    \\ \hline
$q=0.150$ & 1    & 1    & 1    & 1    & 1    & 1    \\ \hline
$q=0.175$ & 1    & 1    & 1    & 1    & 1    & 1    \\ \hline
$q=0.200$ & 1    & 1    & 1    & 1    & 1    & 1    \\ \hline
$q=0.225$ & 1    & 1    & 1    & 1    & 1    & 1    \\ \hline
$q=0.250$ & 0.36 & 1    & 1    & 0.18 & 0.36 & 0.16 \\ \hline
$q=0.275$ & 0.22 & 0.24 & 1    & 0.09 & 0.16 & 0.14 \\ \hline
$q=0.300$ & 0.14 & 0.44 & 0.38 & 0.06 & 0.11 & 0.10 \\ \hline
$q=0.325$ & 0.16 & 0.22 & 0.34 & 0.13 & 0.06 & 0.10 \\ \hline
$q=0.350$ & 0.20 & 0.18 & 0.22 & 0.08 & 0.06 & 0.08 \\ \hline
$q=0.375$ & 0.14 & 0.20 & 0.16 & 0.08 & 0.06 & 0.08 \\ \hline
$q=0.400$ & 0.12 & 0.14 & 0.16 & 0.07 & 0.06 & 0.11 \\ \hline
$q=0.425$ & 0.18 & 0.10 & 0.10 & 0.05 & 0.08 & 0.07 \\ \hline
$q=0.450$ & 0.16 & 0.10 & 0.14 & 0.05 & 0.06 & 0.06 \\ \hline
$q=0.475$ & 0.10 & 0.18 & 0.14 & 0.07 & 0.06 & 0.08 \\ \hline
$q=0.500$ & 0.12 & 0.14 & 0.12 & 0.05 & 0.05 & 0.05 \\ \hline
\end{tabular}
\caption{Table of matching accuracy in the single Erd\H os-R\'enyi background setting with $p=0.5$, averaged over 10 Monte Carlo iterates}
\label{tab:er05}
\end{table}  

\begin{table}[b!]
\centering
\begin{tabular}{|c|c|c|c|c|c|c|}
\hline
$n$         & 50   & 50   & 50   & 100  & 100  & 100  \\ \hline
$m$         & 10   & 100  & 1000 & 10   & 100  & 1000 \\ \hline
$q=0$     & 1    & 1    & 1    & 1    & 1    & 1    \\ \hline
$q=0.025$ & 1    & 1    & 1    & 1    & 1    & 1    \\ \hline
$q=0.050$ & 1    & 1    & 1    & 1    & 1    & 1    \\ \hline
$q=0.075$ & 1    & 1    & 1    & 1    & 1    & 1    \\ \hline
$q=0.100$ & 1    & 1    & 1    & 1    & 1    & 1    \\ \hline
$q=0.125$ & 1    & 1    & 1    & 1    & 1    & 1    \\ \hline
$q=0.150$ & 1    & 1    & 1    & 1    & 1    & 1    \\ \hline
$q=0.175$ & 1    & 1    & 1    & 1    & 1    & 1    \\ \hline
$q=0.200$ & 1    & 1    & 1    & 1    & 1    & 1    \\ \hline
$q=0.225$ & 0.14 &1     & 1    &0.10  & 1    & 1    \\ \hline
$q=0.250$ & 0.40 & 0.50 & 0.42 & 0.12 & 1    & 0.19 \\ \hline
$q=0.275$ & 0.20 & 0.12 & 0.36 & 0.12 & 0.12 & 0.12 \\ \hline
$q=0.300$ & 0.20 & 0.30 & 0.12 & 0.07 & 0.09 & 0.07 \\ \hline
$q=0.325$ & 0.22 & 0.14 & 0.20 & 0.06 & 0.07 & 0.11 \\ \hline
$q=0.350$ & 0.14 & 0.24 & 0.14 & 0.08 & 0.08 & 0.08 \\ \hline
$q=0.375$ & 0.20 & 0.18 & 0.10 & 0.05 & 0.06 & 0.06 \\ \hline
$q=0.400$ & 0.10 & 0.14 & 0.10 & 0.05 & 0.07 & 0.10 \\ \hline
$q=0.425$ & 0.16 & 0.10 & 0.12 & 0.06 & 0.05 & 0.09 \\ \hline
$q=0.450$ & 0.12 & 0.12 & 0.12 & 0.06 & 0.07 & 0.07 \\ \hline
$q=0.475$ & 0.10 & 0.10 & 0.14 & 0.05 & 0.06 & 0.06 \\ \hline
$q=0.500$ & 0.14 & 0.12 & 0.12 & 0.06 & 0.07 & 0.08 \\ \hline
\end{tabular}
\caption{Table of matching accuracy in the single Erd\H os-R\'enyi background setting with $p=1/3$, averaged over 10 Monte Carlo iterates; similar results are obtained in the $p=0.5$ setting; see Appendix \ref{app:addexp} for detail.}
\label{tab:era13}
\vspace{-5mm}
\end{table}  

\begin{figure}[t!]
        \centering
        \includegraphics[width = 0.6\textwidth]{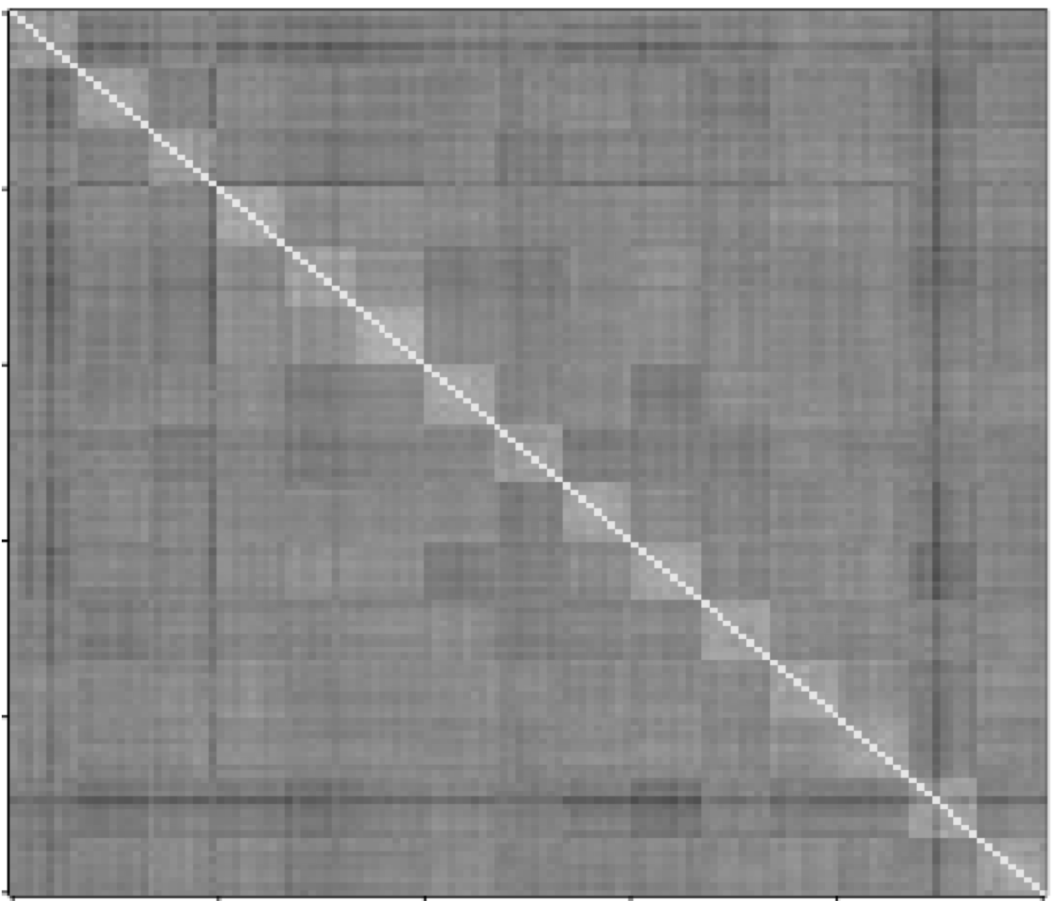}
        \caption{Inter-graph distance matrix heatmap for the 135 in-sample brain graphs considered from the HNU1 dataset, where each subject's scans are plotted contiguously (on the 9$\times$9 diagonal block).
        Larger values in the heatmap are denoted by darker colors.}
        \label{fig:brains3}
    \end{figure}
\subsubsection{Clustering the brain graphs}
\label{app:clust}
To demonstrate how we can obtain the brain graph clusters, we consider the following simple example.  Using 135 in-sample brain graphs considered from the HNU1, we compute the matrix of inter-graph distances $D_{ij}=\|A_i-A_j\|_F$ (displayed in Figure \ref{fig:brains3}).
Embedding this distance matrix into $\mathbb{R}^{14}$ using canonical multidimensional scaling (14 chosen by an elbow analysis of the scree plot of singular values of $D$) and clustering the embedded graphs via $K$-means clustering (with $K=15$, with 25 random restarts) yields an Adjusted Rand Index \cite{rand1971objective} of $1$ (i.e., perfect clustering) between the obtained clusters and the true labels.

\end{document}